\documentclass[10pt,journal,compsoc,twocolumn]{IEEEtran}
\ifCLASSOPTIONcompsoc
  \usepackage[nocompress]{cite}
\else
  \usepackage{cite}
\fi

\ifCLASSINFOpdf
   \usepackage[pdftex]{graphicx}
   \graphicspath{{../pdf/}{../jpeg/}}
   \DeclareGraphicsExtensions{.pdf,.jpeg,.png}
\else
   \usepackage[dvips]{graphicx}
   \graphicspath{{../eps/}}
   \DeclareGraphicsExtensions{.eps}
\fi

\ifCLASSOPTIONcompsoc
 \usepackage[caption=false,font=footnotesize,labelfont=sf,textfont=sf]{subfig}
\else
 \usepackage[caption=false,font=footnotesize]{subfig}
\fi

\usepackage{url}
\usepackage{stfloats}
\usepackage{booktabs}
\usepackage{multirow}
\usepackage[ruled]{algorithm2e}
\usepackage{algpseudocode}
\usepackage{array}
\usepackage{color}
\usepackage{amsmath}
\usepackage{amssymb}
\usepackage{bm}
\newcommand{\UGModel}{UG-AGE}
\newcommand{\DGModel}{DG-AGE}
\newcommand{\HINModel}{HIN-AGE}

\begin{document}
\title{A Robust and Generalized Framework for Adversarial Graph Embedding}

\author{Jianxin~Li,
        Xingcheng~Fu,
        Hao~Peng,
        Senzhang~Wang,
        Shijie~Zhu,
        Qingyun~Sun,
        Philip~S.~Yu,~\IEEEmembership{Fellow,~IEEE,}
        and~Lifang~He,~\IEEEmembership{Member,~IEEE}
\IEEEcompsocitemizethanks{\IEEEcompsocthanksitem Jianxin Li, Xingcheng Fu, Hao Peng, Shijie Zhu and Qingyun Sun are with Beijing Advanced Innovation Center for Big Data and Brain Computing, Beihang University, Beijing 100083, and also with the State Key Laboratory of Software Development Environment, Beihang University, Beijing 100083, China. 
E-mail: \{lijx, fuxc, penghao, zhusj, sunqy\}@act.buaa.edu.cn
\IEEEcompsocthanksitem Senzhang Wang is with the School of Computer Science and Engineering, Central South University, Changsha 410083, China. 
E-mail: szwang@csu.edu.cn.
\IEEEcompsocthanksitem Philip S. Yu is with the Department of Computer Science, University of Illinois at Chicago, Chicago, IL 60607, USA.
E-mail: psyu@uic.edu.
\IEEEcompsocthanksitem Lifang He is with the Department of Computer Science and Engineering, Lehigh University, Bethlehem, PA 18015 USA. E-mail: lih319@lehigh.edu.
}
\thanks{Manuscript received April 2021. (Corresponding author: Hao Peng.)}
}


\IEEEtitleabstractindextext{%
\begin{abstract}
Graph embedding is essential for graph mining tasks. 
With the prevalence of graph data in real-world applications, many methods have been proposed in recent years to learn high quality graph embedding vectors various types of graphs. 
However, most existing methods usually randomly select the negative samples from the original graph to enhance the training data without considering the noise. 
In addition, most of these methods only focus on the explicit graph structures and cannot fully capture complex semantics of edges such as various relationships or asymmetry. 
In order to address these issues, we propose a robust and generalized framework for adversarial graph embedding based on generative adversarial networks. 
Inspired by generative adversarial network, we propose a robust and generalized framework for adversarial graph embedding, named AGE. 
AGE generates the fake neighbor nodes as the enhanced negative samples from the implicit distribution, and enables the discriminator and generator to jointly learn each node’s robust and generalized representation. 
Based on this framework, we propose three models to handle three types of graph data and derive the corresponding optimization algorithms, i.e., \UGModel~ and \DGModel~ for undirected and directed homogeneous graphs, respectively, and \HINModel~for heterogeneous information networks.
Extensive experiments show that our methods consistently and significantly outperform existing state-of-the-art methods across multiple graph mining tasks, including link prediction, node classification, and graph reconstruction. 
\end{abstract}

}

\maketitle

\IEEEpeerreviewmaketitle

\IEEEraisesectionheading{\section{Introduction}\label{sec:introduction}}


\begin{figure*}[htb]
\centering
\hspace{-0.2cm}
\subfloat[Directed graph.]{
\label{fig:example:b}
\begin{minipage}[t]{ 0.45\textwidth}
\centering
\includegraphics[width=\textwidth]{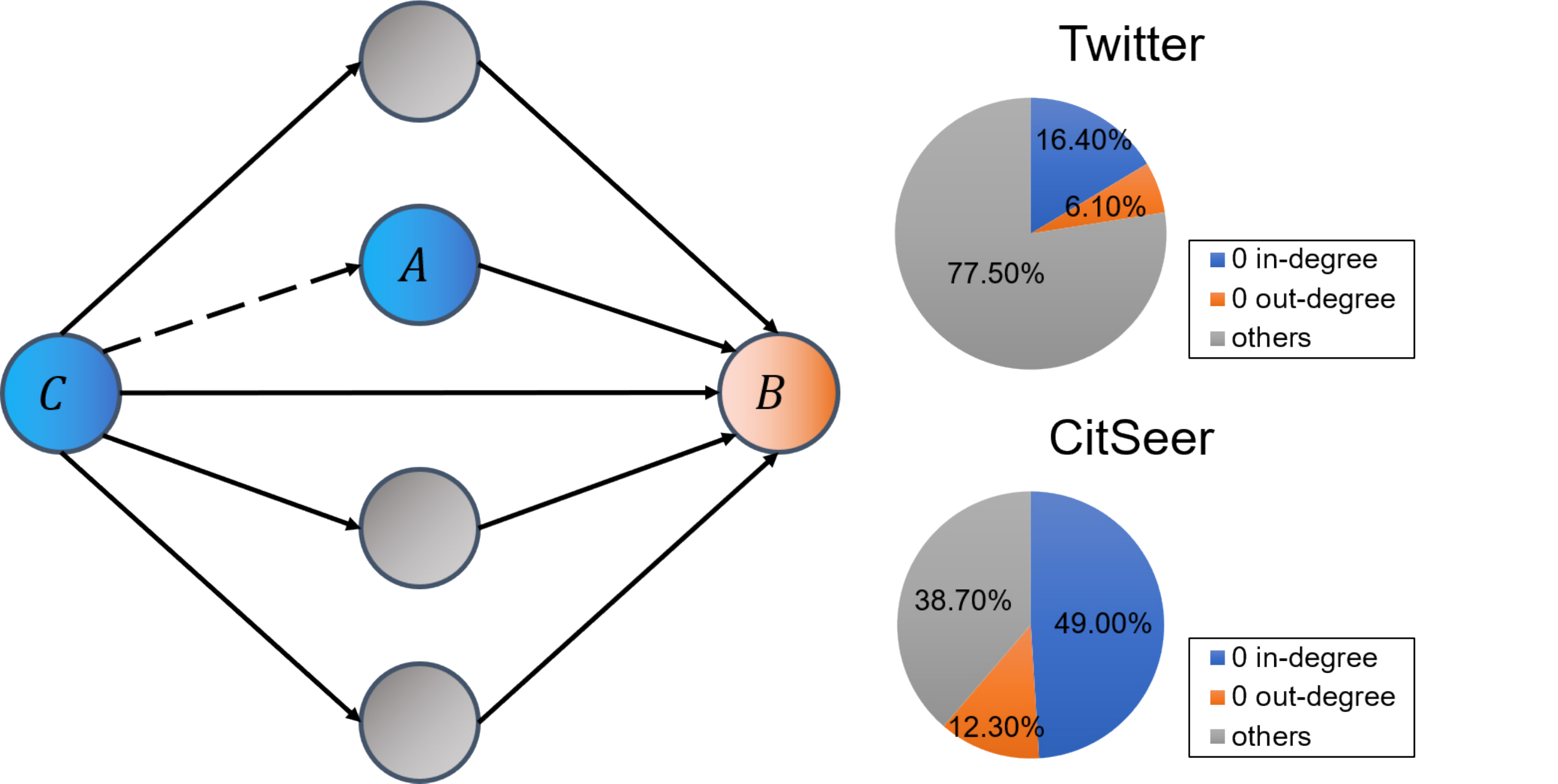}
\end{minipage}
}
\subfloat[heterogeneous information networks.]{
\label{fig:example:a}
\begin{minipage}[t]{ 0.45\textwidth}
\centering
\includegraphics[width=\textwidth]{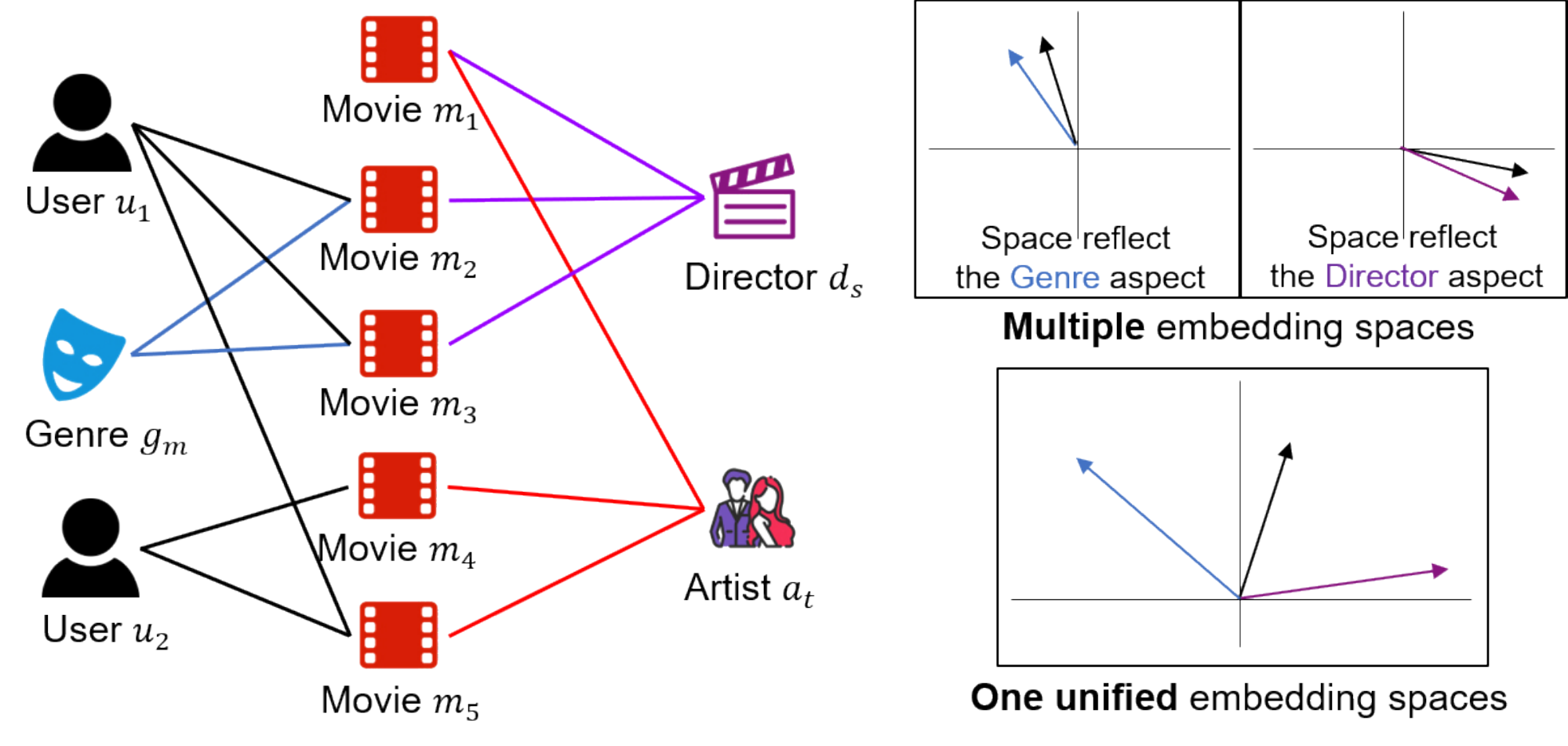}
\end{minipage}
}
\centering
\caption{(a) An example of asymmetric semantics (directed graph). The two pie charts are statistics from the social network of Twitter and the citation network of CiteSeer, respectively. (b) An example of heterogeneous semantics (heterogeneous information networks). On the left sub-figure, different colored lines represent different types of relationships. On the sub-figure, the differences of each node in multiple relationships embedding spaces and unified embedding spaces. }
\label{fig:example}
\end{figure*}

\IEEEPARstart{G}{raph} representation learning aims to learn a low-dimensional vector of each node in a graph, and has gained increasing research attention recently due to its broad applicabilites on graph mining tasks, such as link prediction~\cite{liben2007link}, graph reconstruction~\cite{tsitsulin2018verse}, recommendation~\cite{ying2018graph}, and node classification~\cite{bhagat2011node}. 

Recently, many graph representation learning methods have been proposed for various types of graphs.
These methods can be roughly divided into three types including matrix factorization based methods~\cite{cao2015grarep,wang2017community,ou2016asymmetric}, random walk based methods~\cite{perozzi2014deepwalk,grover2016node2vec,tang2015line,dong2017metapath2vec}, 
and deep learning based methods~\cite{wang2016structural,wu2020comprehensive, kipf2016variational, hamilton2017inductive}. 
Most of these methods rely on strict proximity measures~\cite{katz1953new} and low rank assumption of the graph adjacent matrix, but ignore the existence of noise in the real-world environment and thus lead to the over-fitting problem in the learning process. 
Moreover, most of these methods perform negative sampling from the original graph to speed up and ensure the effect of training. 
These negative samples are limited to the existing samples of graph, and their training strength is not enough, so it is not conducive to improve the generalization of the model. 

Many generative adversarial networks (GAN) based methods~\cite{wang2018graphgan,dai2018adversarial,yu2018learning} have been proposed to solve the above problem by using adversarial training regularization. 
Although these GAN-based methods can learn robust node representations, their generators focus on learning the discrete node connection distribution in the original graph. 
Moreover, since these methods have complex frameworks and require negative sampling or graph softmax function to approximate the node distribution, they are difficult to train and scale to large-scale graphs. 

Meanwhile, many graphs in the real-world contain complex semantics (e.g., social networks, citation networks and web-page networks). 
For graphs with semantics, we argue that there are two major limitations for existing works on improving the robustness of model and preserving semantic information at the same time. 
First, for directed graphs with asymmetry semantic information, existing methods focus on preserving the structure proximities~\cite{zhou2017scalable,katz1953new}  but ignore the underlying semantic information of the nodes. 
For the nodes with only out-degree or in-degree edges, their target or source embedding cannot be effectively trained. 
Fig.~\ref{fig:example:b} presents a toy example of a directed graph. 
For predicting the link between nodes $A$ and $C$ in Fig.~\ref{fig:example:a}, $A$ and $C$ are the nodes with only out-degree edges and $AC$ is a potential link. 
Since the node pair $(A,C)$ is regarded as negative samples, it is hard for existing methods to predict the link $AC$. 
As shown in Fig.~\ref{fig:example:b}, the nodes with zero out-degree or in-degrees (e.g., $A$ and $B$) account for a large proportion of the graph. 
It means these nodes with asymmetric semantic information are ubiquitous in some real-world graphs.
Second, for heterogeneous information networks with different relationships, existing GAN-based methods cannot directly and explicitly model the semantic information of different types of relationships. 
Mapping different types of nodes into a unified low-dimensional space may lead to significant information loss. 
The lack of explicit representation of the graph complex semantics may cause many problems, such as embedding distortion and semantic ambiguity~\cite{shi2018aspem}. 
Fig.~\ref{fig:example:a} shows an example of a film network, where the user $u_1$ has relations with both the musical $g_n$ and the director $d_s$, but $g_n$ and $d_s$ have a low correlation. 
If all the nodes are embedded into a unified low-dimensional space, $u_1$ can only be embedded in the middle of $g_n$ and $d_s$, and make $u_1$ not similar to $g_n$ and $d_s$ anymore. 
Therefore, how to make full use of the complex semantics to improve the graph representation learning becomes extremely challenging. 

To address the above challenges, we propose a novel robust and generalized framework for \underline{A}dversarial \underline{G}raph \underline{E}mbedding (AGE).
Specifically, the generator generates fake neighborhoods for each node from a learned implicit continuous distribution of node representations. 
Competition between the generator and discriminator drives both of them to improve their capability until the generated distribution are indistinguishable from the true connectivity distribution. 
Unlike the exisiting GAN-based methods that need sampling from the original graph, our method generates fake neighbors as negative samples directly from the implicit distribution of each node. 
Our method can more efficiently preserve the semantic information of the graph while being efficient. 
Compared with existing GAN-based methods, our framework is more flexible and scalable, since it generates fake neighbors directly from a continuous distribution and is not sensitive to different graph structures. 
The source code of this work is publicly available at~\texttt{\url{https://github.com/RingBDStack/AGE}}. 

A preliminary version of this work which focuses on adversarial network embedding learning on directed graphs appeared in the proceedings of AAAI 2021~\cite{DGGAN}. 
This journal version has extended it into a unified framework and implements corresponding variant models based on our framework for the representation learning on three typical graphs: undirected graphs, directed graphs, and heterogeneous graphs. 
(1) 
For undirected graphs, we propose a more robust and efficient method named \UGModel, where the adversarial mechanism module acts as a regular term in the graph representation learning process.
The generator samples noise from the implicitly Gaussian distribution of each node, and directly generates fake neighbors as negative samples for adversarial training. 
(2) 
For directed graphs, the challenge is that asymmetric semantic causes the difficulty of learning representations of node with zero out-degrees or in-degrees.
To preserve asymmetric semantic, we propose a asymmetric-aware model named \DGModel, which has two generators for generating fake source neighbors and fake target neighbors, respectively.  
The two generators can enhance each other and they can also learn effective source and target vectors for nodes with zero out-degree or in-degree. 
(3) 
For heterogeneous information networks, the challenge is how to learn the node representation with different relationship semantics. 
To preserve heterogeneity semantic, we propose a relationship-aware model named \HINModel, which can be used with the translate models~\cite{bordes2013translating,wang2014knowledge,Ji2015Knowledge} of knowledge graph embedding with simple modifications to learn the various relationship semantics. 
The generator can directly generate implicit nodes from the continuous distribution for different relationships efficiently and the generated fake neighbors are not limited to existing nodes. 

We integrate above three models together with a robust and generalized framework named AGE.
We conduct extensive experiments on real-world graph datasets, and the results show that the proposed models consistently and significantly outperform various state-of-the-art methods on the tasks of link prediction, node classification and graph reconstruction.

We highlight the advantages of AGE as follows:

$\bullet$ \textbf{Robustness and Generality.} AGE generates adversarial samples from the implicit distribution calculated by the latent node representations. It can be generalized to non-existent nodes and not restricted to the original graph. 

$\bullet$ \textbf{Semantic-preserving.} 
AGE can modify the implicit distribution according to different graph semantics, which can effectively preserve the complex semantics of the graph. 

$\bullet$ \textbf{Scalablity.} 
Since the implicit node distribution is continuous, the proposed framework is scalable to large-scale graphs. 

$\bullet$ \textbf{Flexiblity.} 
Many other graph embedding methods and external knowledge can be plugged into AGE.





The rest of the paper is organized as follows. 
We introduce the overall framework in Section~\ref{sec:Overall Framework}, and propose three variant models for different graphs in Section~\ref{sec:Methods}. 
The experimental results and analysis are presented in Section~\ref{sec:Experiment}. 
We review related work in Section~\ref{sec:Related Work}. 
Finally, conclusion and future work are given in Section~\ref{sec:Conclusion}.

\section{Overall Framework}
\label{sec:Overall Framework}
The proposed framework mainly consists of two components: generator and discriminator, which jointly learn the robust and generalized representations for nodes in the graph. 
The generator learns to generate negative samples by sampling from the implicit distribution of graph and the discriminator learns to distinguish negative samples from the positive ones. 
The overall process is shown in Algorithm~\ref{Overall:algorithm}. 

\subsection{The Implicit Distribution of Graph}
In our framework, the generator first generates false neighbor nodes by sampling from a continuous and uniform noise distribution. 
Then we use the fake nodes as enhanced negative samples to improve the robustness and generalizability of the graph embedding model. 
In order to make the model learn the generalized and potential graph features as much as possible, we design different noise distributions for different graphs by using node implicit representations and other auxiliary information. 
The implicit node distribution based on $d$-dimensional Gaussian distribution $N(\boldsymbol{\mu},\sigma^{2} \mathrm{I})$ can be calculated by node representations as:
\begin{align}
\small
\bm{\eta} \sim N\left(\mathbf{Z}, \sigma^{2} \mathrm{I}\right),
\label{eq1}
\end{align}
where $\bm{\eta}$ is generated noise vector, $\mathbf{Z}$ is the $d$-dimensional implicit node representations, and $\sigma^{2} \mathrm{I} \in R^{d \times d}$ is a covariance variable that can be learned.  
The noise distribution used in our adversarial mechanism is selected based on the implicit feature representation of the graph, which improves the robustness and generalization of the graph embedding model. 

\subsection{Generator}
\noindent
\textbf{Basic generator structure. }
The generator with the implicit node distribution is defined as: 
\begin{align}
\small
G\left(\cdot; \theta^{G}\right)=f\left(\bm{\eta} ; \theta^{f}\right),
\label{eq2}
\end{align}
where $\theta ^{G}$ is the parameters for generator $G$, and the input of $G\left(\cdot; \theta^{G}\right)$ can be the node representations and semantic information of the graph. 
$\theta ^{f}$ is the parameters of the transform function $f$. 
The generator samples the noise from the implicit node distribution $\bm{\eta} \sim N\left(\mathbf{z}_{u}^{T}, \sigma^{2} \mathrm{I}\right)$ according to Eq.~\ref{eq1}, where $\mathbf{z}_{u} \in R^{d \times 1}$ is the embedding vector of node $u$. 
The parameter of generator $G$ is:

\begin{align}
\small
\theta^{G}=\left\{\bm{\eta}: \mathcal{G}, \theta^{f}\right\}. 
\label{eq3}
\end{align}%

Given a node $u$, the generator outputs the embedding $\mathbf{e}_{u^{\prime}} \sim G\left(u:\mathcal{G}, \theta^{G}\right)$ of the false neighbor node $u^{\prime}$. 
In this way, we can obtain a negative sample node pair $(u, u^{\prime})$. 

\begin{figure*}[!t]
\centering
\subfloat[Basic generator]{
\label{fig:generators:a}
\includegraphics[width=0.32\textwidth]{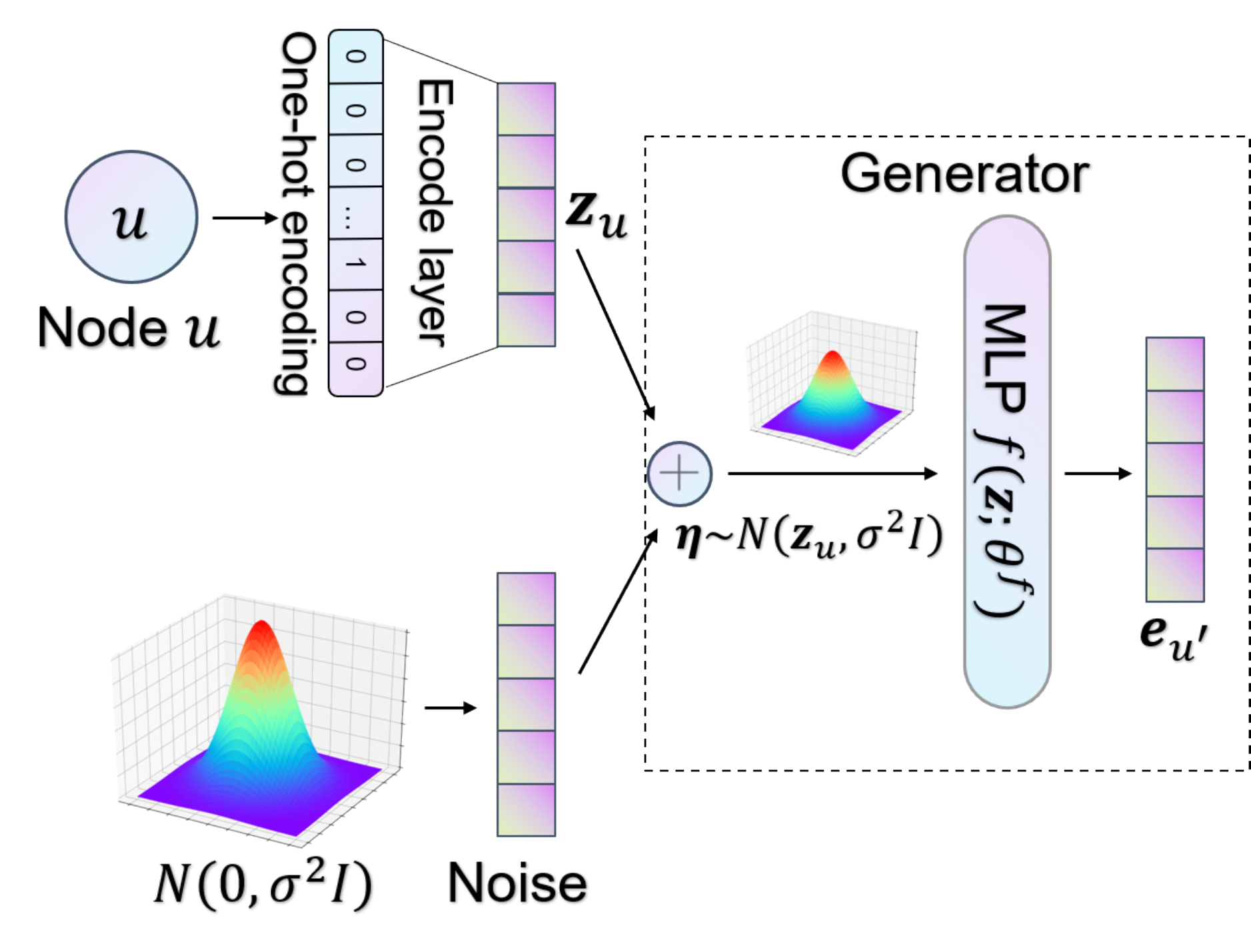}}
\hfil
\hfil
\subfloat[Asymmetry preserved generator]{
\label{fig:generators:c}
\includegraphics[width=0.32\textwidth]{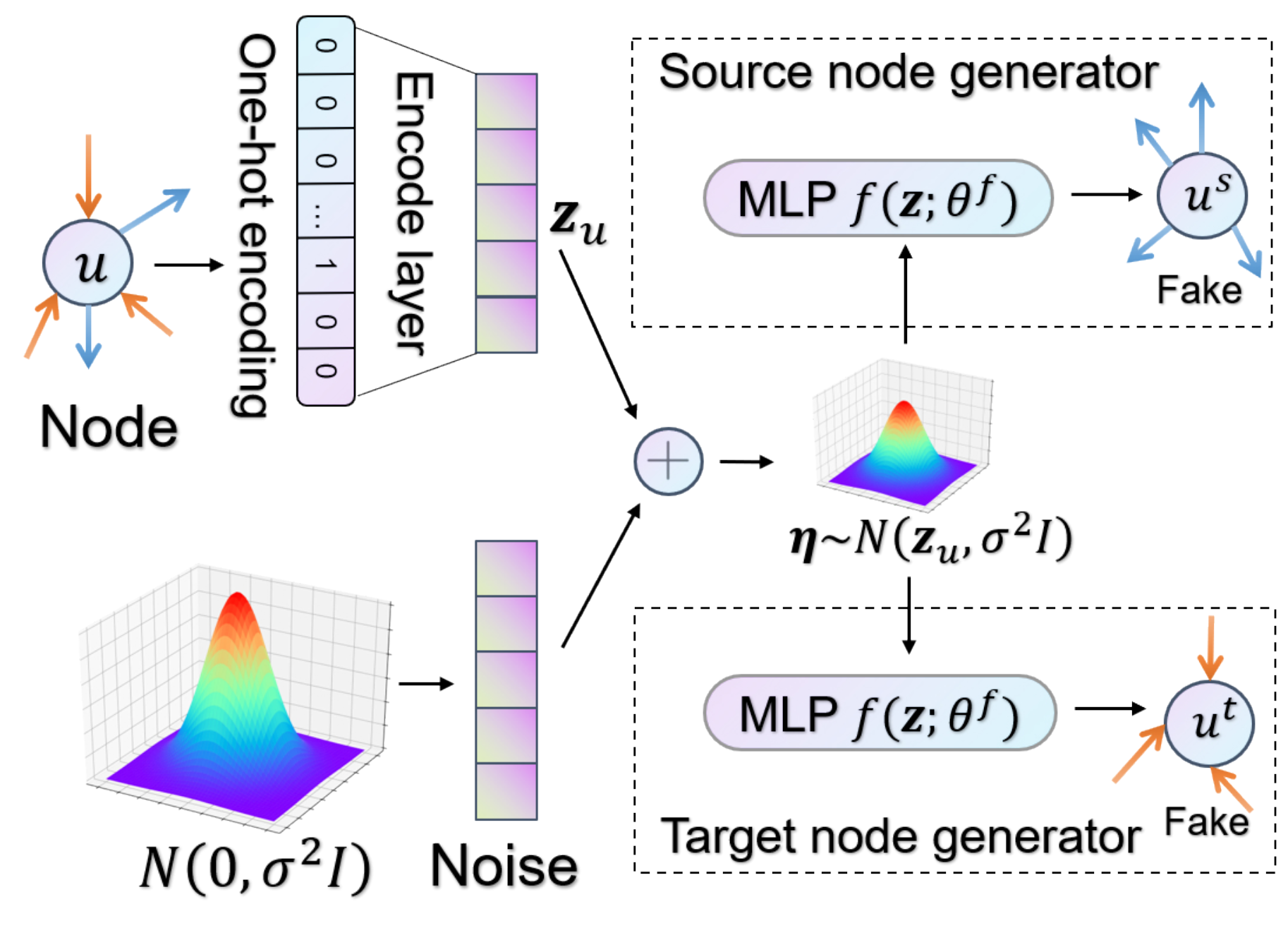}}
\hfil
\hfil
\subfloat[Heterogeneity preserved generator]{
\label{fig:generators:b}
\includegraphics[width=0.32\textwidth]{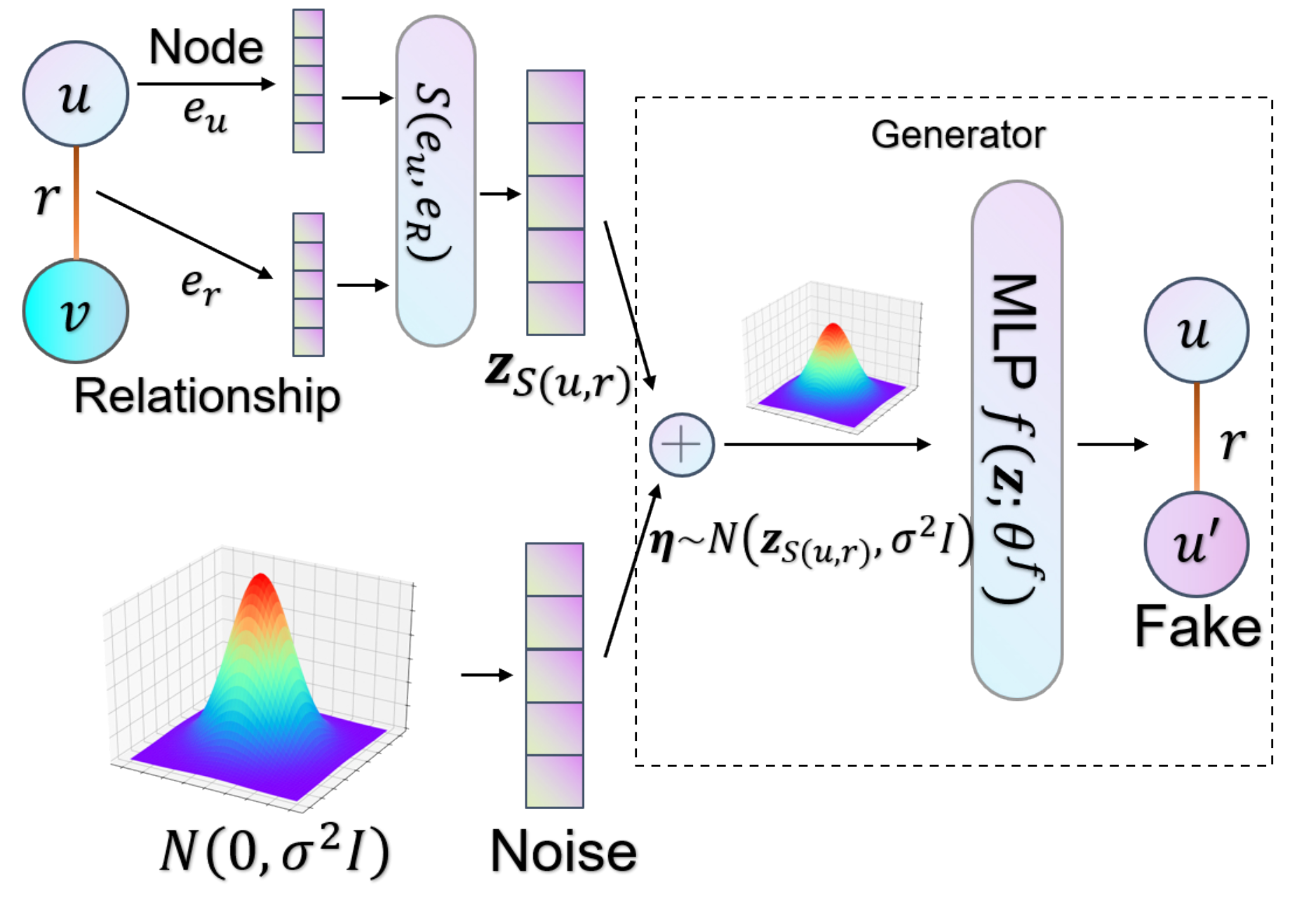}}
\caption{Illustration of the generator of the proposed framework. 
(a) 
The basic generator calculates the implicit node distribution and generates the fake neighbors for each node. 
(b) 
The Asymmetry preserved generator represent source and target properties of nodes with a shared implicit node distribution for graphs with asymmetry semantic. 
(c) 
The Heterogeneity preserved generator fuses representations of nodes and relationships to obtain the implicit node distribution. }
\label{fig:generators}
\end{figure*}

An example of a basic generator implementation is shown in Fig.~\ref{fig:generators:a}. 
First, we get the one-hot encoding of the input node $u$, and then input it to an embedding layer to get a dense vector representation. 
Note that this can be replaced with other embedding models as required. 
At the same time, the generator randomly samples a noise vector from a Gaussian distribution. 
The dense vector and the noise vector are added as the vector $z$, which is used as the input of $f(\cdot)$, and $f(\cdot)$ outputs the embedding of the generated false node $\mathbf{e}_{u^{\prime}}$. 

\noindent
\textbf{Semantics preserved generator. }
For graphs with complex semantics, we need to make full use of their semantic information. 
Therefore, we need to preserve the semantic information of the graph in the generated samples, consistent with the sampling of the positive samples. 
Our solution is to design a distribution that fuses semantics and implicit node representations as the noise distribution of the generator. 
The generator samples random noise from this semantic preserved implicit node distribution and generates fake neighbor nodes for adversarial training. 
For two typical graph semantic information, asymmetry and heterogeneity, we give the semantic-preserved implicit node distribution respectively as follows: 

$\bullet$~\textit{Asymmetry. } 
For the scenarios that are difficult to model asymmetry semantics directly such as directed graphs (DG), we use two generators to learn the source and target representations of the nodes, respectively. 
As shown in Fig.~\ref{fig:generators:c}, the source generator $G^{s}$ and target generator $G^{t}$ share an implicit node distribution: 
\begin{equation}
\begin{aligned}
\small
G^{s}\!(u ; \theta^{G^{s}})=f^{s}\left(\bm{\eta} ; \theta^{f^{s}}\right), 
G^{t}\!(u ; \theta^{G^{t}})=f^{t}\left(\bm{\eta} ; \theta^{f^{t}}\right), 
\end{aligned}
\label{eq6}
\end{equation}
where $\bm{\eta} $ is sampled from the same implicit node distribution. 
By jointly learning, two generators can both capture the source and target semantic information of each node on directed graphs. 

To sum up, as the generator is designed with full consideration of the semantics of a graph and uses continuous implicit distribution to generate fake samples directly, our framework is more adaptive, scalable and computational efficient for different graphs. 

$\bullet$~\textit{Heterogeneity. }
For scenarios that can explicitly model heterogeneous semantics (e.g., heterogeneous information network~\cite{DBLP:journals/tkde/ShiLZSY17}, knowledge graph~\cite{DBLP:journals/tkde/WangMWG17}), we first fuse the node implicit representations and heterogeneous semantic representations as shown in Fig.~\ref{fig:generators:b}. 
Then the noise distribution of the generator can be obtained based on the fused Gaussian distribution. 
Informally, the definition of semantics preserved implicit node distribution can be derived from Eq.~\eqref{eq2} as:
\begin{align}
\small
p(\bm{\eta}| \mathbf{s})=N\left(z_{\mathcal{S}(\mathbf{e}_{u},\mathbf{e}_{r})}, \sigma^{2} \mathrm{I}\right),
\label{eq5}
\end{align}
where $\mathbf{e}_{u}$ and $\mathbf{e}_{r}$ are the node representation of node $u$ and semantics representation of relation $r$, respectively, and $\mathcal{S}(\cdot)$ is a function that fuses the node representation and the semantics representation. 
The generator samples noise from this distribution and generates fake nodes as negative samples. 

\subsection{Discriminator}
The discriminator aims to distinguish strongly connected node pairs from weak ones and calculate the possibility of an edge between the generated node $u^{\prime}$ and the original node $u$. 
For any node pair $(u, v)$, the discriminant function outputs a number from $0$ to $1$ indicating the probability that node pair is true. 
Informally, the discriminator $D$ of our framework is defined as:
\begin{align}
\small
D(u,v;\theta^{D})=F\left(u, v ; \theta^{F}\right),
\label{eq4}
\end{align}
where $\theta^{D}$ is the parameters of discriminator $D$. 
The discriminator can combine with different graph embedding methods, making our framework flexible and easy to extend. 

\begin{algorithm}[t]
\LinesNumbered
\caption{The process of overall framework.} 
\label{Overall:algorithm}
\small
\KwIn{Graph $\mathcal{G}$, the number of maximum training epochs $n^{epoch}$, the numbers of generator and discriminator training iterations per epoch $n^G$, $n^D$, the number of samples $n^s$.}
\KwOut{$\theta^G$, $\theta^D$.}
Initialize $\theta^G$ and $\theta^D$;\\
\For{$epoch = 0; epoch < n^{epoch}$}{
    \For{$n = 0; n < n^D$}{
        \tcp{For each node }
        \For{$u \in \mathcal{V}$}{
            $\bm{\eta} \gets$~Eq.~\eqref{eq1} \tcp*{Fake neighbor}
            $\theta^D \gets$~Eq.~\eqref{eq6} \tcp*{Update discriminator}
        }
    }
    \For{$n = 0; n < n^D$}{
        \tcp{For each node }
        \For{$u \in \mathcal{V}$}{
            $\bm{\eta} \gets$~Eq.~\eqref{eq1} \tcp*{Fake neighbor}
            $\theta^D \gets$~Eq.~\eqref{eq3} \tcp*{Update generator}
        }
    }
}
\end{algorithm} 

\section{Modeling for Different Graphs}
\label{sec:Methods}
In this section, we will introduce the implementation details of our framework on different types of graphs. 
First, we present the \UGModel~model for the undirected homogeneous graph, which represents basic and simple graph without semantics information. 
For the graph with rich semantics, we present \DGModel~and  \HINModel~for two typical type of graphs: the directed graph and the heterogeneous information networks, respectively. 

\subsection{\UGModel: Adversarial Network Embedding for Undirected Graphs}
For undirected homogeneous graphs, we propose \UGModel~based on our framework by a simple modification. 

\noindent
\textbf{Generator of \UGModel. }
According to Eqs.~\eqref{eq1},~\eqref{eq2} and~\eqref{eq3}, we define the generator, implicit distribution, and the generator parameters of \UGModel~ as follows: 
\begin{equation}
\begin{aligned}
\small
G(u, \theta^{G})&=~f(\bm{\eta} ; \theta^{f}),~~
\bm{\eta} \sim N(\mathbf{z}_{u}^{T}, \sigma^{2} \mathrm{I}),\\
\theta^{G}&=~\{\mathbf{z}_{u}: u \in V, \theta^{f}\}.
\end{aligned}
\label{eq7}
\end{equation}

The generator $G$ outputs the embedding $\mathbf{e}_{u^{\prime}} \sim G(u, \theta^{G})$ of the generated false neighbor node $u^{\prime}$. 
In this way, a negative sample node pair $(u, u^{\prime})$ is obtained. 
The generator $G$ is trained to deceive the discriminator and its loss function $\mathcal{L}^{G}$ is:

\begin{align}
\small
\mathcal{L}^{G}=\mathbb{E}_{u \in \mathcal{V}} \log (1-F(u, u^{\prime})),
\label{eq8}
\end{align}
where $F(\cdot )$ is the discriminant function in the discriminator, and its output is a decimal of $0$ to $1$, which represents the probability that the input node pair $(u, u^{\prime})$ is positive. 


\noindent
\textbf{Discriminator of \UGModel. }
The discriminator part is divided into two modules: 
one module is a graph structure reservation module for learning graph structure and the other is an adversarial training module to improve the robustness and generalization of the model. 

$\bullet$~\textit{Graph Structure Preservation Module. }
The purpose of the graph structure preservation module is to preserve the original graph structure in the low-dimensional embedding space. 
Many graph embedding methods can be used directly as the graph structure preservation module, such as DeepWalk~\cite{perozzi2014deepwalk}, LINE~\cite{tang2015line}, node2vec~\cite{grover2016node2vec}, etc. 
Taking DeepWalk as an example, for each node pair $(u, v)$, the corresponding loss function $\mathcal{L}_{\mathrm{NE}}^{D}$ is:
\begin{equation}
\begin{aligned}
\small
\mathcal{L}_{\mathrm{NE}}^{D} = \log\!\sigma(\mathbf{e}_{u}^{\mathrm{T}}\!\cdot\!\mathbf{e}_{v}) + \sum_{k=1}^{K}\!\mathbb{E}_{n \sim p_{k}(u)}\!\log\!\sigma\!(-\mathbf{e}_{u}^{\mathrm{T}}\!\cdot \!\mathbf{e}_{n}\!),
\label{eq9}
\end{aligned}
\end{equation}
where $\sigma(\cdot)$ is the sigmoid function, $\mathit{K}$ is the number of negative samples, and $p_{k}(u)$ is the sampling distribution of negative samples (usually $p_{k}(u)=d_{v}^{3 / 4} / \sum_{v \in V} d_{v}^{3 / 4}$).
Note that the node pair $(u, v)$ comes from the random walk sampling adopted by DeepWalk and can be modified appropriately according to the specific method. 

$\bullet$~\textit{Adversarial Training of Module. }
The purpose of the adversarial training module is to judge the authenticity of the input node pair. 
For the input node pair $(u,v)$, we use the sigmoid function as the discriminant function $F(u,v;\theta^{F})$ which outputs the probability that the node pair is true. 
For node $u$, the generator $G$ generates a fake neighbor node $u^{\prime}$ and obtains the node pair $(u, u^{\prime})$. 
The loss function with the discriminant function $F$ as an adversarial training module can be obtained by Eq.~\eqref{eq11}:
\begin{align}
\small
\mathcal{L}_{\mathrm{adv}}^{D}=\mathbb{E}_{u \in V}-\log (1-F(u, u^{\prime})).
\label{eq11}
\end{align}%
Considering the graph structure retention module and the adversarial training module, the loss function of the discriminator $D$ is: 
\begin{align}
\small
\mathcal{L}^{D}=\mathcal{L}_{\mathrm{NE}}^{D}+\lambda \mathcal{L}_{\mathrm{adv}}^{D},
\label{eq12}
\end{align}
where $\lambda > 0$ is the weight of the adversarial training module.

\begin{figure*}[!t]
\centering
\subfloat[\UGModel.]{
\includegraphics[width=0.32\textwidth]
{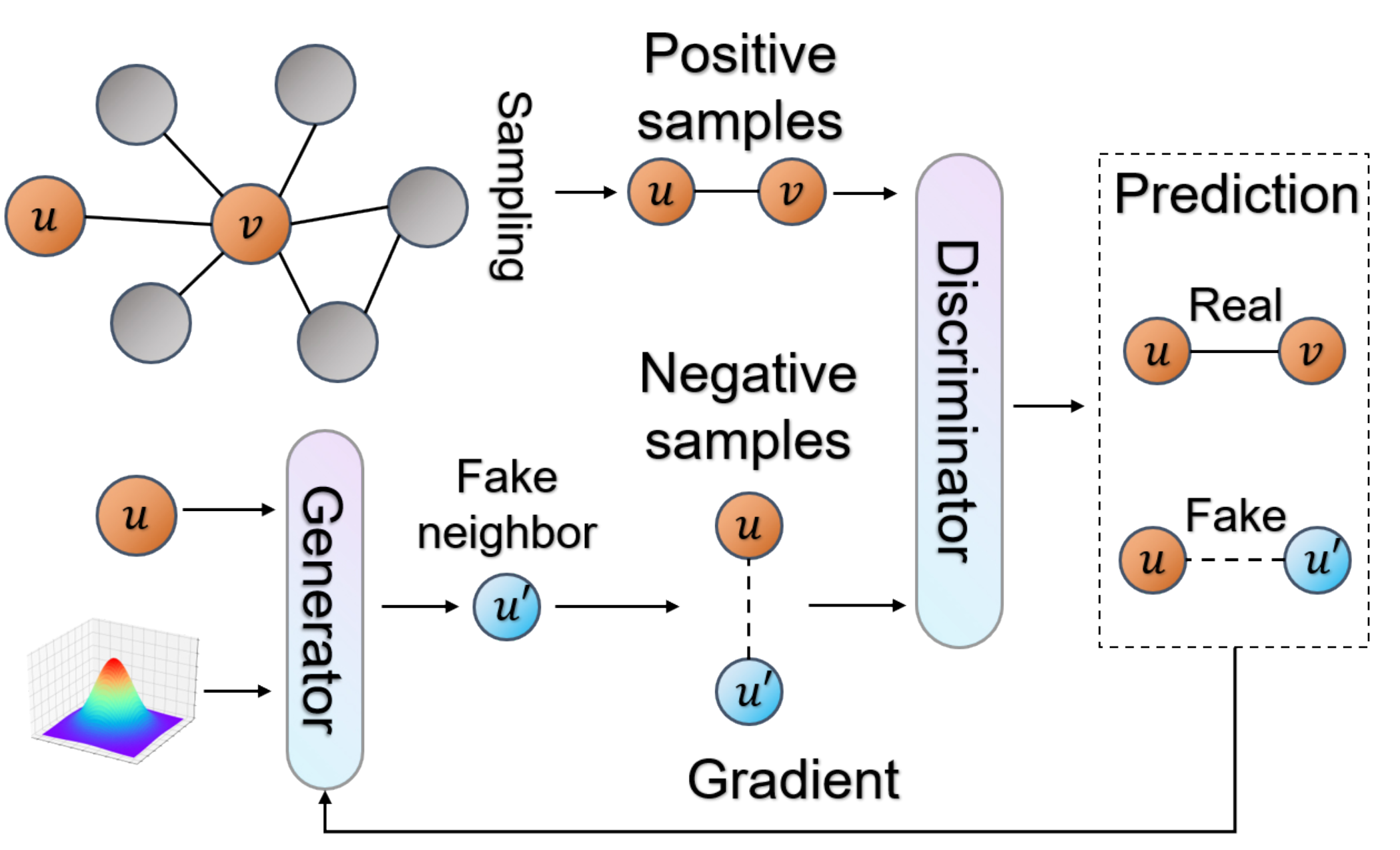}
\label{fig:architecture:a}
}
\hfil
\hfil
\subfloat[\DGModel.]{
\includegraphics[width=0.32\textwidth]{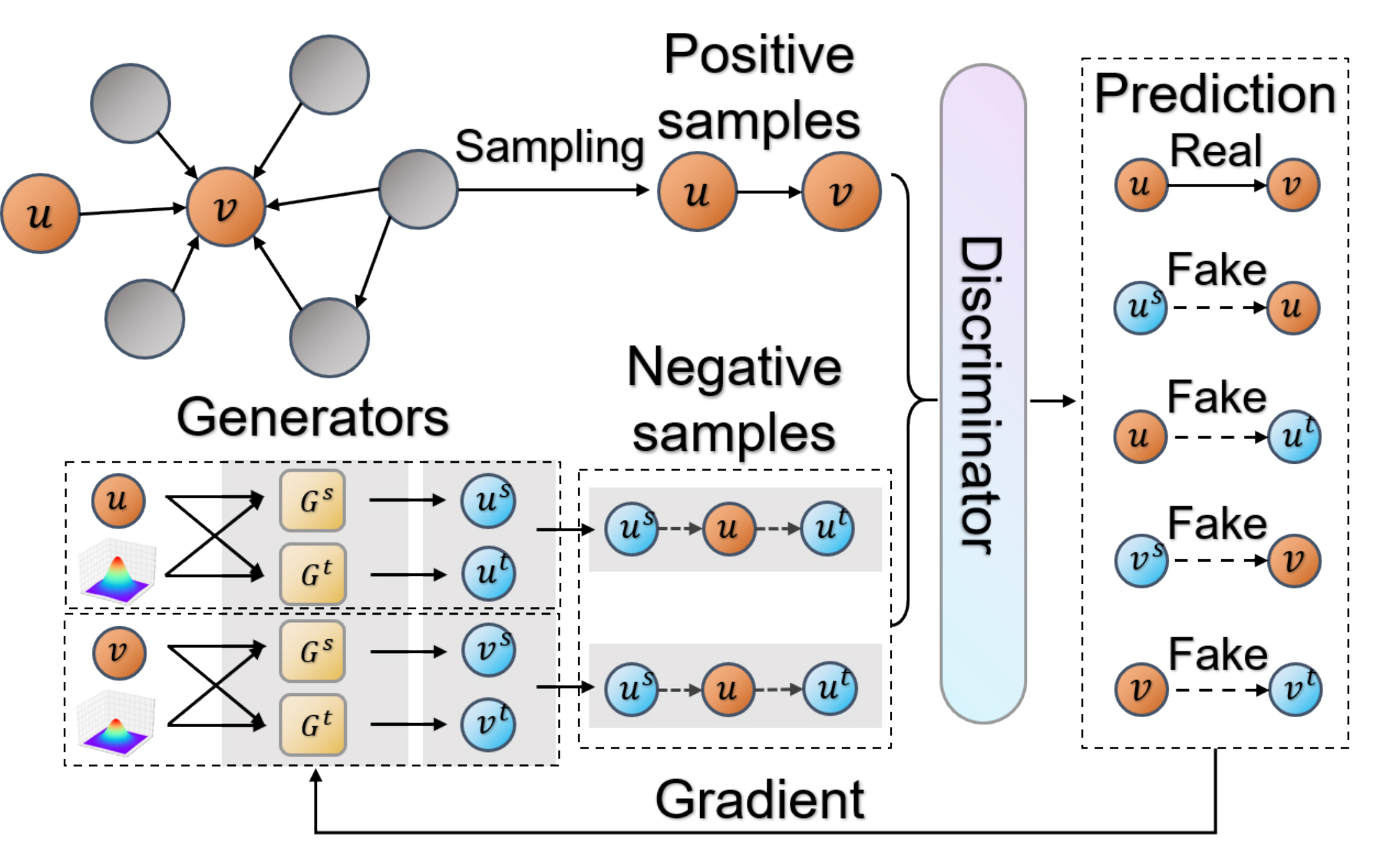}
\label{fig:architecture:b}
}
\hfil
\hfil
\subfloat[\HINModel.]{
\includegraphics[width=0.32\textwidth]
{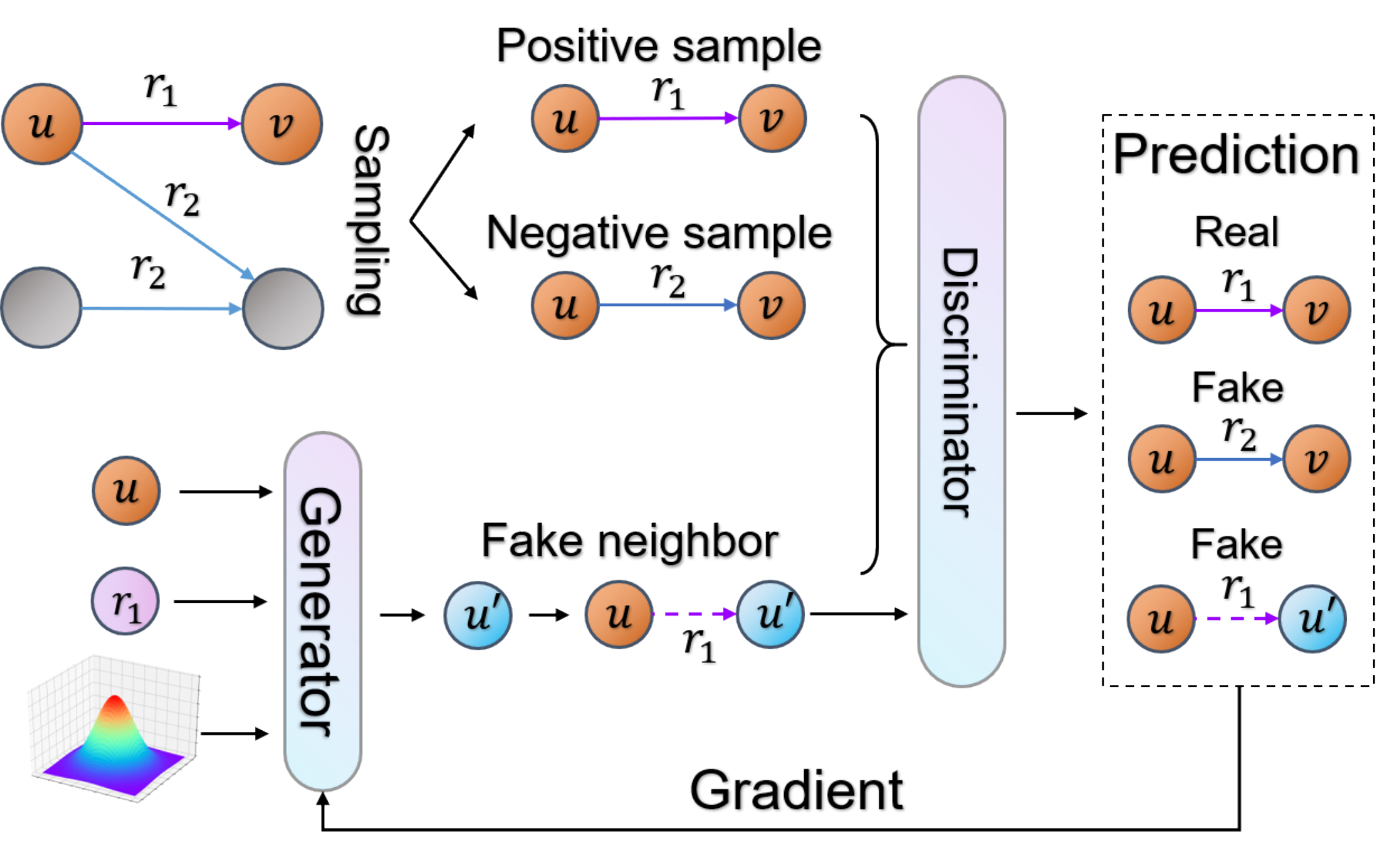}
\label{fig:architecture:c}
}
\caption{Illustration of \UGModel, \DGModel~and \HINModel. (a) \UGModel: for each node, the generator generates fake neighbors as negative samples. (b) \DGModel: for each node, the two generators share an implicit distribution and jointly generate a fake source neighbor and a fake target neighbor as the negative samples. (c) \HINModel: for nodes with different relationships, the generator generate false neighbor nodes by the implicit distribution of corresponding relationships. }
\label{fig:architecture}
\end{figure*}

\noindent
\textbf{Model Optimization. }
As shown in Fig.~\ref{fig:architecture:a}, the model training process is as follows. 
First, to obtain a node pair as a positive sample, a node $u\in \mathcal{V}$ and its neighbor node $v\in \mathcal{V} $ are selected by random walk. 
For node $u$, the generator generates a negative sample $u^{\prime}$ and the negative node pair $(u, u^{\prime})$ is input into the discriminant function $F(\cdot )$.
Second, the loss function $\mathcal{L}^{G}$ is calculated according to the discriminant result of the discriminant function. 
Finally, we update the parameters of the generator according to $\mathcal{L}^{G}$. 
Then we repeat these steps to train the generator and discriminator alternatively until convergence. 

\subsection{\DGModel: Adversarial Graph Embedding for Directed Graphs}
For directed graphs, the key is how to explicitly learn the asymmetric semantics of the graphs. 
To preserve the asymmetric proximity, each node $u$ of a directed graph $\mathcal{G}$ needs to have two different representations according to two roles (i.e., the source role and target role), represented by two $d$ dimensional vector $\mathbf{s}_u \in \mathbb{R}^{d \times 1}$ and $\mathbf{t}_u \in \mathbb{R}^{d \times 1}$, respectively. 

A critical problem arises that nodes with low in-degree or low out-degree are often difficult to learn due to the edges' asymmetry. 
To address this problem, we propose \DGModel~to learn more robust source and target vectors for those nodes with low in-degree or low out-degree, even for nodes with zero in-degree or zero out-degree (such as $u$ and $v$ in Fig.~\ref{fig:architecture:c}). 

\noindent
\textbf{Asymmetry-Aware Generator. }
The generator $G$ has three main goals:
(1) 
$G$ should generate corresponding fake samples under a specific direction. 
Therefore, given a node $u \in \mathcal{V}$, the generator $G$ aims to generate a false source neighbor $u^{s}$ and a false target neighbor $u^{t}$. 
$u^{s}$ and $u^{t}$ should be as close as possible to the real neighbor nodes. 
(2) 
$G$ should generalize well to non-existent nodes. 
In other words, the fake nodes $u^{s}$ and $u^{t}$ cannot be limited to the original graph. 
(3) 
For those nodes with relatively low or zero in-degree or out-degree, $G$ should also be able to generate effective false source neighbors and target neighbors. 

In order to achieve the first goal, the generator $G$ in \DGModel~is composed of two generators: 
the source neighbor generator $G^{s}$ and the target neighbor generator $G^{t}$. 
For the second and third goals, \DGModel~introduces an implicit variable (noised embedding) $\bm{\eta}$ shared between $G^{s}$ and $G^{t}$ to generate negative samples. 
\DGModel~applies two transform functions $f^{s}$ and $f^{t}$ to the generators to enhance the expression ability of fake samples rather than directly generating samples from the implicit distribution. 
Therefore, the formula of generator $G$ is defined as follows:
\begin{align}
\small
G(u; \theta^{G})&=\{G^{s}(u; \theta^{G^{s}}), G^{t}(u ; \theta^{G^{t}})\}, 
\label{eq19}
\end{align}
where $\theta^{f^{s}}$ and $\theta^{f^{t}}$ represent the parameters of $f^{s}$ and $f^{t}$, respectively. 
The noised embedding $\bm{\eta}$ serves as a bridge between $G^{s}$ and $G^{t}$. 
With the help of $\bm{\eta}$, $G^{s}$ and $G^{t}$ update collaboratively to generate better false source neighbors and target neighbors. 
According to Eq.~\eqref{eq1}, we derive $\bm{\eta}$ from the implicit distribution $\bm{\eta} \sim N(\mathbf{z}_{u}^{\mathrm{T}}, \sigma^{2} \mathbf{I})$, where $\mathbf{z}_{u} \in \mathbb{R}^{d \times 1}$ is a learnable variable, representing the implicit representation of $u \in \mathcal{V}$. 
The parameters of $G^{s}$ and $G^{t}$ are $\theta^{G^{s}}=\{\mathbf{z}_{u}^{\mathrm{T}}: u \in \mathcal{V}, \theta^{f^{s}}\}$, $\theta^{G^{t}}=\{\mathbf{z}_{u}^{\mathrm{T}}: u \in \mathcal{V}, \theta^{f^{t}}\}$. 

The two generators $G^{s}$ and $G^{t}$ aim to deceive the discriminator $D$ by generating fake samples close to real ones. 
Therefore, the loss function $\mathcal{L}^{G}$ of generators $G^{s}$ and $G^{t}$ is defined as follows:
\begin{align}
\small
\mathcal{L}^{G}=\mathbb{E}_{u \in v} (\log (1-D(u^{s}, u))+\log (1-D(u, u^{t}))),
\label{eq22}
\end{align}%
where $u^{s}$ and $u^{t}$ represent the false source neighbors of node $u$. 
The source vector $\mathbf{s}_{u^{s}}$ and the target vector $\mathbf{t}_{u^{t}}$ of node $u$ can obtained by $\mathbf{s}_{u^{s}} \sim G^{s}(u ; \theta^{G^{s}})$, and $\mathbf{t}_{u^{t}} \sim G^{t}(u ; \theta^{G^{t}})$. 
The parameters of $G^{s}$ and $G^{t}$ can be optimized by minimizing $\mathcal{L}^{G}$. 

\noindent
\textbf{Asymmetry-Aware Discriminator. }
The discriminator $D$ aims to distinguish the negative samples generated by the generator $G$ from the positive samples input sampled from the original graph $\mathcal{G}$. 
Note that for a given node pair $(u, v)$, $D$ outputs the probability that the node $v$ is connected to the node $u$ in the out-degree direction. 
In particular, the input node pair can be divided two cases as follows:

$\bullet$~\textit{Positive Sample.} 
There is indeed a directed edge from $u$ to $v$ on $\mathcal{G}$ (i.e., $(u, v) \in \mathcal{E}$). 
In this case, the node pair $(u, v)$ is considered to be positive and the corresponding loss function is: 
\begin{align}
\small
\mathcal{L}_{\mathrm{pos}}^{D}=\mathbb{E}_{(u, v) \sim p_{ \mathcal{G} } }-\log D(u, v). 
\label{eq24}
\end{align}


$\bullet$~\textit{Negative sample.}
Given node $u \in \mathcal{V}$, $u^{s}$ and $u^{t}$ represent its false source neighbors and false target neighbors, which are generated by $G^{s}$ and $G^{t}$, respectively. 
In this case, node pairs such as $(u^{s}, u)$ and $(u, u^{t})$ are considered to be negative and the loss function is: 
\begin{equation}
\begin{aligned}
\small
\mathcal{L}_{\text {neg }}^{D}=~\mathbb{E}_{u \in v}-\log (\!1\!-\!D(u^{s}, u)\!) - \log(\!1\!-\!D(u, u^{t})\!). 
\label{eq25}
\end{aligned}
\end{equation}

Note that the discriminator $D$ treats the fake node representations $\mathbf{s}_{u^{s}}$ and $\mathbf{t}_{u^{t}}$ as unlearnable inputs. 
Integrating the above two cases together, the loss function $\mathcal{L}^{D}$ of the discriminator $D$:
\begin{align}
\small
\mathcal{L}^{D}=\mathcal{L}_{\text {pos }}^{D}+\mathcal{L}_{\text {neg }}^{D}.  
\label{eq26}
\end{align}
The discriminator $D$ can be optimized by minimizing $\mathcal{L}^{D}$. 

\noindent
\textbf{Model Optimization of \DGModel. }
In each training epoch, \DGModel~uses mini-batch gradient descent to train the discriminator $D$ and the generator $G$ alternatively. 
Specifically, \DGModel~first fixes $\theta^{G}$ and generates corresponding fake neighbors for each node pairs of the graph to optimize $\theta^{D}$. 
Then, \DGModel~fixes $\theta^{D}$ and each node generates false neighbor nodes close to the real ones to optimize $\theta^{G}$ under the guidance of $D$. 
The generator and discriminator conduct adversarial training until \DGModel~converges. 

\begin{table}[t]
\caption{Noise Distribution of \HINModel~Generator.}
\scriptsize
\resizebox{0.48\textwidth}{!}{
\newcommand{\tabincell}[2]{\begin{tabular}{@{}#1@{}}#2\end{tabular}}
\centering
\begin{tabular}{|c|c|c|}
\hline 
Model &Noise Distribution &Auxiliary Operation \\
\hline 
TransE &$N(\mathbf{e}_{u}^{G}+\mathbf{e}_{r}^{G}, \sigma^{2} \mathbf{I})$ &-\\
\hline 
TransH &$N(\mathbf{e}_{u,r}^{G}+\mathbf{e}_{r}^{G}, \sigma^{2} \mathbf{I})$  &$\mathbf{e}_{u,r}^{G} =\mathbf{e}_{u}^{G}-\mathbf{w}_{r}^{{G}^{T} }\mathbf{e}_{u}^{G}\mathbf{w}_{r}^{G}$ \\
\hline 
TransD &$N(\mathbf{e}_{u}^{G}\mathbf{M}_{r,u}^{G}  +\mathbf{e}_{r}^{G}, \sigma^{2} \mathbf{I})$  &\tabincell{c}{$\mathbf{M}_{r,u}^{G}=\mathbf{r}_{p}^{G}\mathbf{u}_{p}^{{G}^{T} }+\mathbf{I}$
\\
$\mathbf{M}_{r,v}^{G}=\mathbf{r}_{p}^{G}\mathbf{u}_{p}^{{G}^{T} }+\mathbf{I}$}  \\
\hline 
\end{tabular}
}
\label{\HINModel:Generator}
\end{table}

\subsection{\HINModel: Adversarial Graph Embedding for Heterogeneous Information Networks}
For heterogeneous information networks, an essential problem is how to explicitly model the various relationship semantics of the graph. 
To preserve different relationship semantics, given a node $u\in \mathcal{V} $ and a relation $r\in \mathcal{R} $, we need to generate a fake node $u^{\prime}$ that may be connected to $u$ with a relationship $r$ in the context. 

Inspired by the translate model of knowledge graph, we generate fake neighbors for edges with different relationship semantics by using relationship embedding. 

\noindent
\textbf{Relationship-Aware Generator. }
The generator $G(\cdot;\theta^{G})$ has two main goals. 
First, $G$ can generate negative nodes close to the real sample. 
Second, $G$ must be relationship-aware and the generated fake neighbor $u^{\prime}$ should be as close to the real node as possible under this relationship. 

In order to meet the above requirements, the implicit distribution of \HINModel's generator is closely related to the graph relations. 
We design \HINModel~model based on three commonly used translate models in the knowledge graph: TransE~\cite{bordes2013translating},  TransH~\cite{wang2014knowledge} and TransD~\cite{Ji2015Knowledge}. 
Specifically, according to different translate models, the generator uses the corresponding Gaussian distribution, as shown in Table~\ref{\HINModel:Generator}. 
Taking the TransE method as an example, the noise distribution $N(\mathbf{e}_{u}^{G}+\mathbf{e}_{r}^{G}, \sigma^{2} \mathbf{I})$ is a Gaussian distribution with mean value $\mathbf{e} _{u}^{G}+\mathbf{e} _{r}^{G} $ and covariance $\sigma^{2} \mathbf{I} \in \mathbb{R}^{d \times d}$. 
Intuitively, the mean value represents the representation vector of fake nodes that may be connected to $u$ through the relationship $r$, and the covariance represents the potential deviation. 
For generators, \HINModel~uses a transform function $f$ instead of direct sampling for higher quality negative samples, and the formula of generator can be defined as follows: 
\begin{align}
\small
G(u,r; \theta^{G})=f(\bm{\eta}; \theta^{f})
\label{eq13},
\end{align}%
where $\bm{\eta} \sim N(\mathbf{e}_{u}^{G}+\mathbf{e}_{r}^{G}, \sigma^{2} \mathbf{I})$, $\theta ^{f}$ is the parameters of $f$ and 
$\theta^{G}=\{\mathbf{e}_{u}^{G}: u \in \mathcal{V}, \mathbf{e}_{r}^{G}: r \in \mathcal{R}, \theta^{f}\}$ is the  parameters of $G$. 
The \HINModel~generator structure is shown in Fig.~\ref{fig:generators:b} and the loss function $\mathcal{L}^{G}$ of the generator $G$ is defined as follow: 
\begin{equation}
\begin{aligned}
\small
\mathcal{L}^{G}=~ \mathbb{E}_{(u, r) \sim \operatorname{p}_{\mathcal{G} }, \boldsymbol{\theta}_{u^{\prime}} \sim G(u, r ; \theta^{G})} -\log (1-D(\mathbf{e}_{u^{\prime}} \mid u, r)). 
\label{eq14}
\end{aligned}
\end{equation}
$\theta^{G}$ can be optimized by minimizing $\mathcal{L} ^{G}$.

\begin{table}[t]
\caption{Score Function of \HINModel~Discriminator.}
\newcommand{\tabincell}[2]{\begin{tabular}{@{}#1@{}}#2\end{tabular}}
\centering
\resizebox{0.48\textwidth}{!}{
\begin{tabular}{|c|c|c|}
\hline 
Model &Score Function &Auxiliary Operation \\
\hline 
TransE &${ \| \mathbf{e}_{u}^{D}+\mathbf{e}_{r}^{D}-\mathbf{e}_{v}^{D}  \| } _{L_1/L_2}$ &-\\
\hline 
TransH &${ \| \mathbf{e}_{u,r}^{D}+\mathbf{e}_{r}^{D}-\mathbf{e}_{v,r}^{D}  \| } _{L_1/L_2}
$  &$\mathbf{e}_{u,r}^{D} =\mathbf{e}_{u}^{D}-\mathbf{w}_{r}^{{D}^{T} }\mathbf{e}_{u}^{D}\mathbf{w}_{r}^{D}$ \\
\hline 
TransD &${ \| \mathbf{e}_{u}^{D}\mathbf{M}_{r,u}^{D}+\mathbf{e}_{r}^{D}-\mathbf{e}_{v}^{D}\mathbf{M}_{r,v}^{D}  \| } _{L_1/L_2}$  &\tabincell{c}{$\mathbf{M}_{r,u}^{D}=\mathbf{r}_{p}^{D}\mathbf{u}_{p}^{{D}^{T} }+\mathbf{I}$
\\
$\mathbf{M}_{r,v}^{D}=\mathbf{r}_{p}^{D}\mathbf{u}_{p}^{{D}^{T} }+\mathbf{I}$}  \\
\hline 
\end{tabular}
}
\label{\HINModel:Discriminator}
\end{table}

\noindent
\textbf{Relationship-Aware Discriminator. }
For heterogeneous information networks, the discriminator aims to distinguish between real and fake nodes under a given relationship. 
Specifically, given a heterogeneous information networks $\mathcal{G}$ and a relation $r$, the discriminator $D(\mathbf{e}_{v} \mid u, r ; \theta^{D})$ outputs the probability that sample $v$ is connected to $u$ under $r$. 
This probability can be quantified as the score function in translate models as shown in Table~\ref{\HINModel:Discriminator}.
Given a node $u$ and a relation $r$, sample a node $v$ and each triple $(u, r, v)$ belongs to one of the following three cases: 

$\bullet$~\textit{Real node connected under a real relation: $(u,r,v)$. }
The nodes $u$ and $v$ are actually connected under the relation $r$ on the heterogeneous information networks $\mathcal{G}$. 
Such triples can be modeled by the following loss function:
\begin{align}
\small
\mathcal{L}_{1}^{D}=\mathbb{E}_{(u, v, r) \sim p_{\mathcal{G}}}-\log D(\mathbf{e}_{v}^{D} \mid u, r). 
\label{eq15}
\end{align}
In this case, the triple $(u,r,v)\sim p_{\mathcal{G} }$ is sampled from $\mathcal{G}$ and the discriminator should judge them as positive. 

$\bullet$~\textit{Real node connected under a fake relation: $(u,r',v)$. }
A node pair $u$ and $v$ are connected in the fake relationship $r^{\prime} \ne r$.
The discriminator should judge them as negative because their connection relation does not match the given relation $r$. 
The loss function of this case is:
\begin{align}
\small
\mathcal{L}_{2}^{D}=\mathbb{E}_{(u, v) \sim p_{g}, r^{\prime} \sim p_{R^{\prime}}}-\log (1-D(\mathbf{e}_{v}^{D} \mid u, r^{\prime})). 
\label{eq16}
\end{align}
In this case, a pair of nodes $(u,v)$ is sampled from $\mathcal{G}$ and the fake relation $r^{\prime}$ is generated by uniformly sampling from ${\mathcal{R}}' = \mathcal{R}\setminus  \{r\} $. 

$\bullet$~\textit{Fake node under a real relation: $(u,r,v')$. }
Given a node $u\in \mathcal{V}$ and a relation $r$, the generator $G(u,r;\theta ^{G} )$ generates a fake neighbor $v'$ for $u$ under the relation $r$. 
Similarly, the discriminator should judge this triple as negative, and the loss function is as follows:
\begin{equation}
\begin{aligned}
\small
\mathcal{L}_{3}^{D}=\mathbb{E}_{(u, r)\sim\!p_{\mathcal{G}},e_{v^{\prime}}\sim\!G(u, r; \theta^{G})} -\log (1\!-\!D(\mathbf{e}_{v^{\prime}}\!\mid\!u, r)).  
\label{eq17}
\end{aligned}
\end{equation}%
Note that the embedding $\mathbf{e} _{v'}$ of fake neighbor $v'$ is sampled from the distribution learned by the generator $G$. 
The discriminator $D$ just treats $\mathbf{e} _{v'} $ as an unlearnable input and only optimizes its own parameters $\theta ^{D}$. 

Note that there is a fourth case where the triples are fake node pairs that are irrelevant and generated by the generator. 
However, this case are easier to distinguish and is implicit in other cases, so we only consider the above three cases and integrate their loss functions to train the discriminator:
\begin{align}
\small
\mathcal{L}^{D}=\mathcal{L}_{1}^{D}+\mathcal{L}_{2}^{D}+\mathcal{L}_{3}^{D}. 
\label{eq18}
\end{align}
The parameters $\theta ^{D}$ of the discriminator can be optimized by minimizing $\mathcal{L}^{D}$. 

\noindent
\textbf{Model Optimization of \HINModel. }
We adopt an iterative optimization strategy to train \HINModel. 
In each iteration, the generator and the discriminator are alternately trained. 
Specifically, we first fix $\theta ^{G}$ and generate fake samples to optimize $\theta ^{D}$ to train the discriminator. 
Then, we fix $\theta ^{D}$ and optimize $\theta ^{G}$ to generate better fake samples. 
Repeat the above process for some iterations until the model converges. 


\subsection{Model Complexity Analysis}
In this section, we analyze the time complexity and space complexity of the adversarial training module in the proposed three models (i.e., \UGModel, \DGModel, and \HINModel). 
For the three model based on our framework, $\mathcal{G}=(\mathcal{V},\mathcal{E})$ is the input graph, $n^s$ is the number of samples, $n^G$ and $n^D$ are the numbers of training iterations of the generator and discriminator respectively, and $d$ is the dimension of the node embedding vectors. 
The detailed explanation and analysis for our models are shown in Table~\ref{tab:complexity}. 

\renewcommand{\multirowsetup}{\centering}
\begin{table}[t!]
\caption{Adversarial training complexity of \UGModel, \DGModel~and \HINModel.}
\scriptsize
\centering
\resizebox{0.5\textwidth}{!}{
\begin{tabular}{@{}c|cc@{}}
\toprule
Model  & Time Complexity & Space Complexity \\ \midrule
\UGModel~ & $O(n^{s} \cdot(n^{D}+n^{G}) \cdot|\mathcal{V}| \cdot d^{2})$ & $O( d \cdot|\mathcal{V}|+|\mathcal{E}|)$ \\
\DGModel~ & $O(n^{s} \cdot(n^{D} \cdot|\mathcal{E}|+n^{G} \cdot|\mathcal{V}|) \cdot d^{2})$ & $O(2 \cdot d \cdot|\mathcal{V}|+|\mathcal{E}|)$ \\
\HINModel~ &$O(n^{s} \cdot(n^{D}+n^{G}) \cdot|\mathcal{E}| \cdot d^{2})$ & $O( d \cdot|\mathcal{V}|+(d+1) \cdot|\mathcal{E}|)$ \\
\bottomrule
\end{tabular}
}
\label{tab:complexity}
\end{table}
\noindent
\textbf{\UGModel.}
As shown in Table~\ref{tab:complexity}, the time complexity and space complexity of adversarial training in \UGModel~are linear to $\mathcal{V}$ (the number of nodes) in the graph. 
Since \UGModel~depends on the specific graph embedding method, the adversarial learning module does not significantly increase the overall model complexity. 

\noindent
\textbf{\DGModel.}
For a directed graph, \DGModel~uses two generators to generate fake neighbors for all nodes and a discriminator to discriminate for all edges. 
Therefore, the time complexity is a linear combination of $|\mathcal{V}|$ and $|\mathcal{E}|$, and the space complexity includes twice the size of the node embedding. 

\noindent
\textbf{\HINModel.}
For heterogeneous information networks, \HINModel~needs to generate fake neighbors based on different types of edges (relationships), so all triples need to be enumerated in the worst case.
For this reason, the time complexity and space complexity of \HINModel~depends not only on the number of nodes $|\mathcal{V}|$, but also on the number of edges $|\mathcal{E}|$.
Overall, the overall computational complexity is linear to both $|\mathcal{V}|$ and $|\mathcal{E}|$. 

Although the three models differ in time complexity and space complexity, the overall complexity of the adversarial training module is linear to the number of nodes and edges of the graph.
In conclusion, the adversarial training module of our framework is both time and space efficient and is scalable for large scale graphs.  

\section{Experiment}
\label{sec:Experiment}
In this section, we conduct extensive experiments on several datasets to investigate the performance of \UGModel, \DGModel~and \HINModel, respectively.

\renewcommand{\multirowsetup}{\centering}
\begin{table}[htbp]
\caption{Statistics of the datasets. (UG: homogeneous undirected graph; \\
DG: directed graph; HIN: heterogeneous information networks.)}
\centering
\resizebox{0.5\textwidth}{!}{
\begin{tabular}{@{}c|ccccc@{}}
\toprule
Dataset  & \#Nodes & \#Edges & Avg. degree  & \#Node classes & \#Graph type \\ \midrule
Cora     & 2,708   & 5,278   & 3.90    & 7    & UG/DG        \\
Citeseer  & 3,264   & 4,551   & 2.79   & 6    & UG        \\
Facebook & 6,637   & 249,967 & 37.66   & 3    & UG        \\
CoCit    & 44,034  & 195,361 & 8.86    & 15   & DG       \\
Twitter  & 465,017 & 834,797 & 3.59    & -    & DG        \\
Epinions & 75,879  & 508,837 & 13.41   & -    & DG        \\
Google   & 15,763  & 171,206 & 21.72   & -    & DG        \\ 
DBLP     & 37,791  & 170,794 & 9.04    & 4    & HIN        \\
Yelp     & 3,913   & 38,680  & 19.77   & 3    & HIN        \\
Aminer   & 312,776 & 599,951 & 3.84    & 6    & HIN        \\
\bottomrule
\end{tabular}
}
\label{dataset1}
\end{table}



\subsection{Datasets and Experiment Setting}
We evaluate the proposed framework on three types of graphs including undirected and directed homogeneous networks, heterogeneous information networks. 
The statistics of these datasets are summarized in Table~\ref{dataset1}.




\noindent
\textbf{Undirected networks. }
\textit{Cora}~\cite{Cora}, \textit{Citeseer}~\cite{mccallum2000automating} are citation networks of academic papers, where nodes are papers, edges are the citation relationships between papers, and labels is the conferences in which papers are published.
\textit{Facebook}~\cite{Facebook} is a social network where nodes are users and edges are the relationships between users.

\noindent
\textbf{Directed networks. }
For citation networks \textit{Cora}~\cite{Cora}  and \textit{CoCit}~\cite{tsitsulin2018verse}, nodes represent papers and directed edges represent the citation relationships between papers. Labels represent conferences in which papers are published.
For social network \textit{Twitter}~\cite{Twitter}, nodes represent users and directed edges represent following relationships between users.
For trust network \textit{Epinions}~\cite{Epinions}, nodes represent users and directed edges represent trust between users.
For hyperlink network \textit{Google}~\cite{Google}, nodes represent pages and directed edges represent hyperlink between pages.

\noindent
\textbf{Heterogeneous information networks. }
\textit{DBLP}~\cite{fu2017hin2vec} and \textit{Aminer}~\cite{HeGAN2019} are scholar networks where nodes are papers, authors and venues, edges are authorships and papers’ venues. 
\textit{Yelp}~\cite{fu2017hin2vec} is a social network where node types include users, businesses, cities, and categories, and edge types include user-user, users’ reviews, business-city, and businesses’ categories.


The parameter settings of all baselines follow the settings in the original model. 
The number of walks, walk length and window size are set to 10, 80 and 10 for comparison. 
node2vec is optimized with grid search over its return and in-out parameters $(p, q) \in \{0.25, 0.50, 1, 2, 4\}$ on each dataset and task. 
For each proposed model, we choose parameters by cross validation and fix the numbers of generator and discriminator training iterations per epoch $n^G=5, n^D=15$ across all datasets and tasks.
Throughout our experiments, the default setting of the dimension of node embeddings is 128. 
The reported results are the average performance of 10 times experiments.

\renewcommand{\multirowsetup}{\centering} 
\begin{table*}[!t]
\setlength{\abovecaptionskip}{0.cm}
\setlength{\belowcaptionskip}{-0.9cm}
\caption{Summary of link prediction AUC scores (\%), node classification Micro-F1 and Macro-F1 scores (\%) on undirected homogeneous graphs. \\
(Result: average score ± standard deviation; \textbf{Bold}: best; \underline{Underline}: runner-up.)}
\centering
\scriptsize
{

\begin{tabular}{c|ccc|ccc|ccc}
\toprule
\multirow{3}{*}{Method} & \multicolumn{3}{c}{Cora}                                     & \multicolumn{3}{c}{Citeseer}                                  & \multicolumn{3}{c}{Facebook}                                                                \\ \cline{2-10} 
                        & AUC            & Micro-F1         & Macro-F1          & AUC            & Micro-F1         & Macro-F1               & AUC            & Micro-F1         & Macro-F1           \\ 
\midrule
DeepWalk        & 80.6±0.12   & 77.8±0.39   & 76.4±0.66   & 73.6±0.31   & 50.1±0.08  & 46.8±0.92   & 85.2±0.37   & 80.3±0.68    & 78.2±0.04       \\
LINE1            & 73.8±0.30   & 78.0±0.61   & 76.5±0.16   & 72.4±0.30   & 52.1±0.34   & 47.5±0.22   & 82.3±0.84   & 79.6±0.22    & 75.6±0.43       \\
node2vec        & 83.8±0.09   & 78.4±1.48   & 77.1±0.90   & 77.6±0.10   & 53.8±1.28   & 50.1±0.70   & 85.5±0.43   & 81.2±0.56    & 79.3±0.54       \\
\midrule
GraphGAN        & 82.5±0.64   & 76.4±0.21   & 76.8±0.34   & 74.5±0.02   & 49.8±1.02   & 45.7±0.13   & 84.2±0.23   & 78.5±1.33    & 75.2±0.96       \\
ANE             & 83.1±0.57   & 78.5±0.51   & 77.0±1.40   & 75.0±1.20   & 50.2±0.12   & 49.5±0.61   & 85.6±1.35   & 82.1±0.14    & 79.6±0.38     \\
\midrule
\UGModel-DW        & \underline{92.58±0.05}   & \underline{83.6±0.06}   & \underline{82.7±0.07}  & \underline{89.81±0.06}   & \underline{63.5±1.29}   & \underline{59.8±0.32}  & \underline{87.5±0.03} & \underline{84.5±0.38}  & \underline{82.6±0.09}      \\
\UGModel-NV        & \textbf{92.62±0.20}   & \textbf{83.9±0.57}   & \textbf{83.1±0.09}  & \textbf{90.28±0.01}   & \textbf{64.1±1.02}   & \textbf{60.5±0.28}  & \textbf{88.3±0.96}   & \textbf{85.6±0.04}   & \textbf{83.5±0.72}    \\ 
\bottomrule
\end{tabular}}
\label{NEGAN}
\end{table*}

\subsection{Baselines and Evaluation Metrics}
To evaluate the proposed \UGModel, \DGModel~and \HINModel, we compare them with several graph embedding methods.

\noindent
\textbf{Traditional graph embedding methods. }
DeepWalk~\cite{perozzi2014deepwalk} uses local information obtained from truncated random walks to learn node embeddings.
LINE~\cite{tang2015line} learns large-scale network embedding using first-order and second-order proximities namely LINE-1 and LINE-2, respectively. 
node2vec~\cite{grover2016node2vec} 
uses a biased random walk algorithm for efficiently exploring the neighborhood architecture. 

\noindent
\textbf{GAN-based graph embedding methods. }
GraphGAN~\cite{wang2018graphgan} generates the sampling distribution to sample negative nodes from the graph.
ANE~\cite{dai2018adversarial} trains a discriminator to push the embedding distribution to match the fixed prior.

\noindent
\textbf{Directed graph embedding methods. }
HOPE~\cite{ou2016asymmetric} preserves the asymmetric information of the nodes by approximating high-order proximity. 
APP~\cite{zhou2017scalable} proposes a random walk based method to encode Rooted PageRank proximity. 

\noindent
\textbf{Heterogeneous information networks embedding methods. }
HIN2vec~\cite{fu2017hin2vec} retains the semantics of heterogeneous information networkss by learning nodes and meta-paths simultaneously. 
Metapath2vec~\cite{dong2017metapath2vec} samples random walks based on meta-paths to preserve semantics. 
TransE~\cite{bordes2013translating} optimizes the distance of sampled triples in a heterogeneous information networks.

For the link prediction (LP) task, the evaluation metric is the area under curve (AUC) score of the ROC. 
For the node classification (NC) task, the evaluation metrics are the Micro-F1 score and Macro-F1 score. 
For the graph reconstruction task, the evaluation metric is Precision score. 

\subsection{Performance Evaluation of \UGModel}



For the evaluation of \UGModel, we compare it with several methods, including traditional graph embedding methods and GAN-based graph embedding methods. 
We also propose two versions of \UGModel~implemented by using DeepWalk and node2vec for network structure retention, named \UGModel-DW and \UGModel-NV, respectively. 

\noindent
\textbf{Performance Analysis.}
To evaluate the performance of \UGModel, we perform two tasks, link prediction and node classification.
For the link prediction task, we predict missing edges given a graph with a fraction of removed edges. 
Specifically, we remove 20\% of edges on \textit{Cora}, \textit{Citseer} and \textit{Facebook} as positive samples, and randomly select node pairs with unconnected edges as negative samples in the test set. 
Note that we make sure that no node is isolated to avoid meaningless embedding vectors when randomly removing edges.
The ratio of training to test data is 8:2. 
For the node classification task, we evaluate the proposed \UGModel~and baseline methods on three undirected homogeneous graph dataset \textit{Cora}, \textit{Citseer} and \textit{Facebook}. 
We randomly sample a fraction of the labeled nodes as the training data and train a standard one-vs-rest $L_2$-norm regularized logistic regression classifier. 
Then we predict the labels of the other nodes. 

We show AUC scores, Micro-F1 and Macro-F1 of all models in Table~\ref{NEGAN}. 
We can notice that compared with DeepWalk and node2vec, \UGModel-DW and \UGModel-NV with the adversarial training module can achieve higher AUC scores and F1 scores, which verifies the adversarial training module is considerably beneficial to preserve the graph structure. 
NEGAN-DW and \UGModel-NV both perform better than DeepWalk, node2vec and LINE on the three datasets. 
The underlying reason is that these baselines generate negative samples by randomly sampling from the original graph, which are not strong enough and can be easily identified by the model, while the negative samples of \UGModel-DW and \UGModel-NV are generated by implicit node distributions. 
Through adversarial learning between the generator and the discriminator, \UGModel~improves model's generalization ability. 
Compared with GraphGAN and ANE, our \UGModel-DW and \UGModel-NV also show better performance. 

\begin{figure}[!t]
\centering
\subfloat[Sparsity.]{
\label{fig:NEGAN-GR:a}
    \begin{minipage}{0.48\linewidth}
    \centering
    \includegraphics[width=1\linewidth]{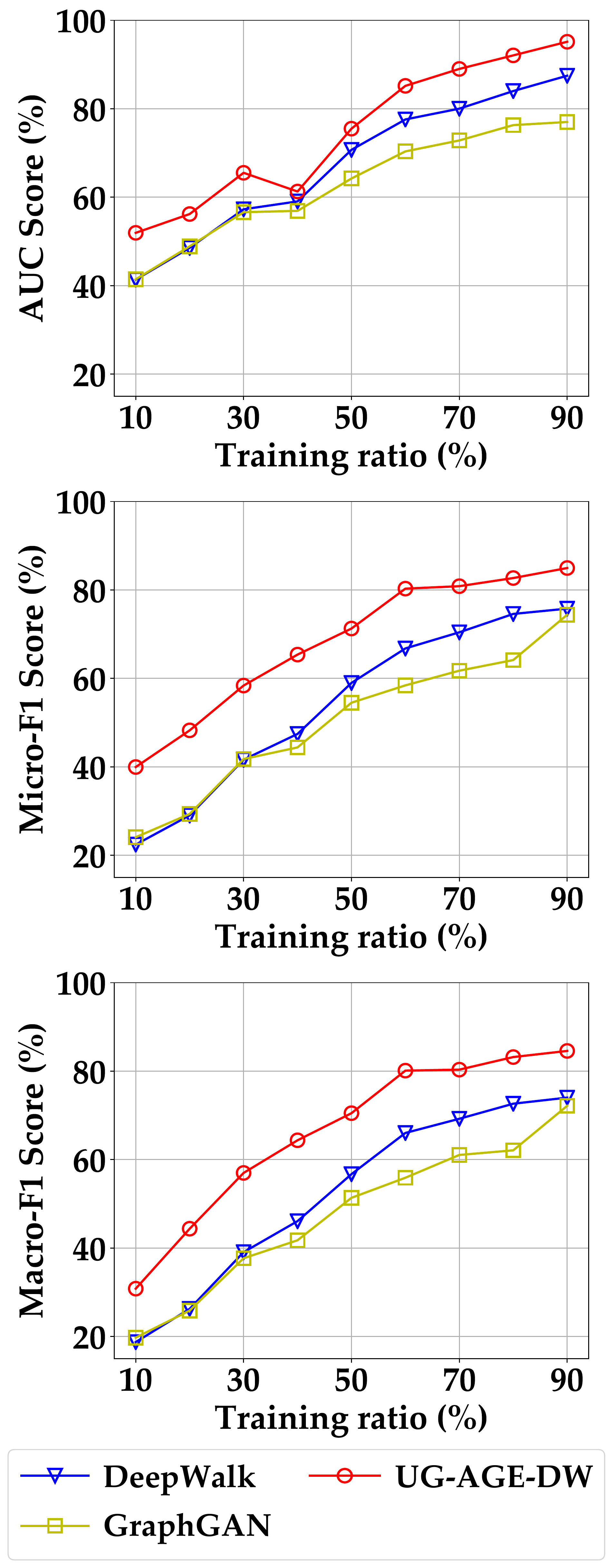}
    \end{minipage}%
}%
\subfloat[Learning curves.]{
\label{fig:NEGAN-GR:b}
    \begin{minipage}{0.48\linewidth}
    \centering
    \includegraphics[width=1\linewidth]{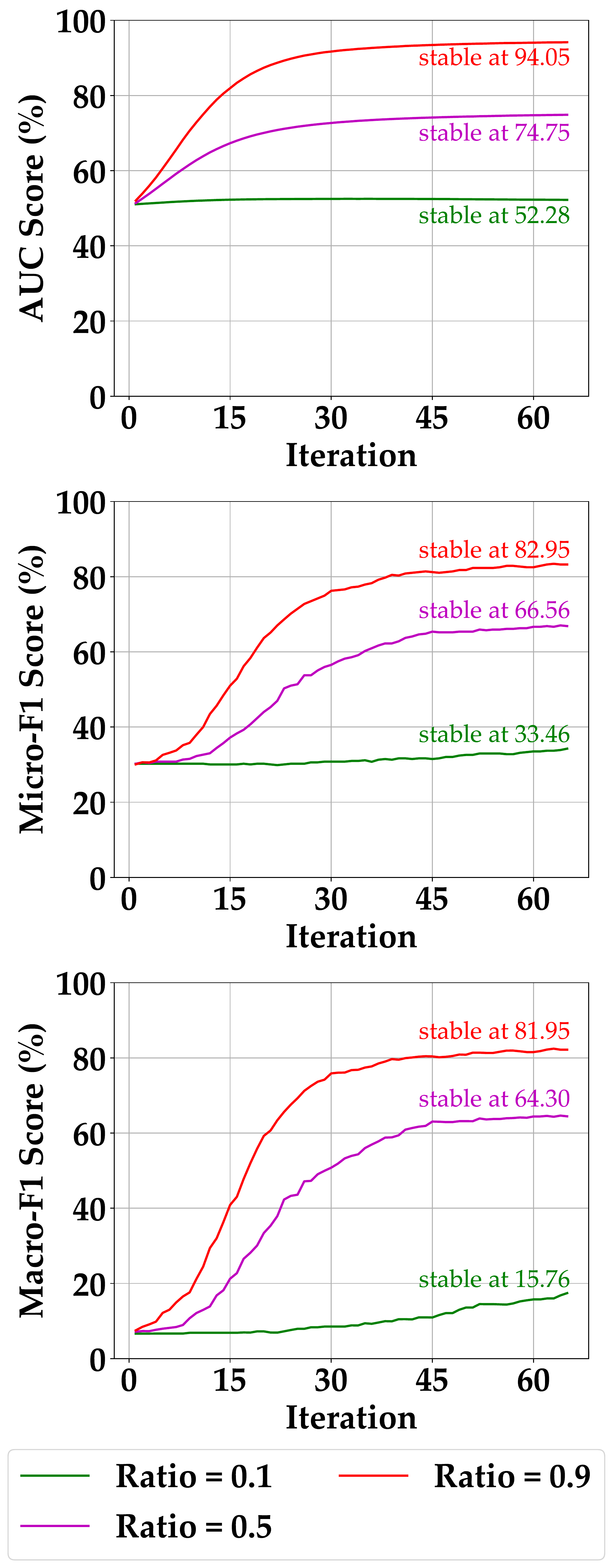}
    \end{minipage}%
}%
\caption{Performance and learning curves of \UGModel~on \textit{Cora} for link prediction and node classification.}
\label{fig:NEGAN-GR}
\end{figure}


\noindent
\textbf{Sparsity and Learning Analysis.}
To verify the advantages of our model, we analyze the link prediction and node classification performance of \UGModel~under different conditions on \textit{Cora}. 
We randomly sample nodes of different ratio (from 10\% to 90\%) as the training data and randomly sample 10\% nodes outside the training set as the test data. 

Fig.~\ref{fig:NEGAN-GR} illustrates the performance and learning curves of \UGModel~with different training ratios on \textit{Cora}. 
Fig.~\ref{fig:NEGAN-GR:a} shows that \UGModel~outperforms baselines under all training ratios, which indicates that \UGModel~adversarial training module can significantly improve the performance of the model. 
Fig.~\ref{fig:NEGAN-GR:b} shows the test performances of \UGModel~in the learning process. 
The results show that the performance of \UGModel~improves rapidly with the increase of the training ratio, indicates that \UGModel~has better generalization ability, especially in node classification. 

\subsection{Performance Evaluation of \DGModel}
For directed graph, we compare \DGModel~with traditional graph embedding methods, directed graph embedding methods, and GAN-based graph embedding methods. 
We also construct a variant of \DGModel, named \DGModel*, which uses only one generator $G^t$ to generate target neighborhoods of each node. 
Note that we do not report the results of GraphGAN on \textit{Twitter} and \textit{Epinions}, since it cannot run on these two large datasets.

\begin{table*}[htbp]
\setlength{\abovecaptionskip}{0.cm}
\setlength{\belowcaptionskip}{-0.9cm}
\caption{Summary of link prediction AUC scores (\%) on directed graphs with various fractions of reversed positive edges. \\ (Result: average score ± standard deviation; \textbf{Bold}: best; \underline{Underline}: runner-up.)
}
\centering
\resizebox{\textwidth}{!}{
\begin{tabular}{c|ccc|ccc|ccc|ccc}
\toprule
\multirow{2}{*}{Method} & \multicolumn{3}{c}{Cora}                                     & \multicolumn{3}{c}{Twitter}                                  & \multicolumn{3}{c}{Epinions}                                 & \multicolumn{3}{c}{Google}                                   \\ \cline{2-13} 
                        & 0\%                & 50\%               & 100\%              & 0\%                & 50\%               & 100\%              & 0\%                & 50\%               & 100\%              & 0\%                & 50\%               & 100\%              \\ 
\midrule
DeepWalk                & 84.9±1.39 & 68.1±0.43          & 52.9±0.12          & 50.4±0.67          & 50.3±0.21          & 50.3±0.01          & 76.6±1.21          & 67.9±0.54          & 66.6±0.12          & 83.6±2.61          & 72.1±0.65          & 66.5±0.32          \\
LINE-1                  & 84.7±0.63          & 68.0±0.25          & 52.5±0.06          & 53.1±0.45          & 51.5±0.13          & 50.0±0.01          & 78.8±0.52          & 69.8±0.26          & 68.5±0.05          & 89.7±0.82          & 72.7±0.45          & 65.1±0.21          \\
node2vec                & \textbf{85.3±1.07}          & 65.5±0.35          & 52.1±0.09          & 50.6±0.75          & 50.5±0.33          & 50.3±0.01          & 89.7±0.31          & 74.6±0.12          & 72.6±0.02          & 84.3±1.13          & 70.5±0.53          & 64.3±0.26          \\ 
\midrule
GraphGAN                & 51.6±0.67          & 51.3±0.31          & 51.2±0.12          & -                  & -                  & -                  & -                  & -                  & -                  & 71.3±2.37          & 61.1±1.59          & 56.2±1.13          \\
ANE                     & 72.8±0.53          & 61.4±0.28          & 51.5±0.07          & 49.7±0.53          & 49.8±0.29          & 50.0±0.02          & 85.5±2.15          & 69.2±0.74          & 66.9±0.24          & 76.1±1.86          & 63.7±0.83          & 57.8±0.53          \\ 
\midrule
LINE-2                  & 69.3±0.47          & 72.1±0.23          & 74.3±0.05          & 95.6±0.37          & 95.7±0.13          & 95.8±0.01          & 58.1±0.67          & 67.1±0.52          & 68.4±0.41          & 77.4±0.24          & 85.2±0.17          & \underline{89.0±0.13}    \\
HOPE                    & 77.6±1.53          & 74.2±0.65          & 71.5±0.42          & 98.0±0.63          & 97.9±0.42          & 97.8±0.03          & 79.6±1.13          & 71.7±0.57          & 70.5±0.23          & 87.5±0.46          & 85.6±0.32          & 84.6±0.38          \\
APP                     & 76.6±0.83          & 76.4±0.41          & 76.2±0.11          & 71.6±0.57          & 70.1±0.36          & 68.7±0.01          & 70.5±0.47          & 71.3±0.23          & 71.4±0.09          & \underline{92.1±0.21}    & 86.4±0.15          & 81.0±0.13          \\ 
\midrule
\DGModel*                  & 83.0±0.91          & \underline{83.3±0.53}    & \underline{83.5±0.25}    & \underline{99.4±0.27}    & \underline{99.3±0.12}    & \underline{99.2±0.01}    & \underline{92.7±0.85}    & \underline{80.0±0.36}    & \underline{78.2±0.21}    & 91.6±0.63          & \underline{89.2±0.47}    & 87.7±0.26          \\
\DGModel                   & \underline{85.1±0.63}    & \textbf{86.7±0.31} & \textbf{88.3±0.11} & \textbf{99.7±0.15} & \textbf{99.7±0.09} & \textbf{99.7±0.01} & \textbf{96.1±0.51} & \textbf{86.4±0.25} & \textbf{85.1±0.11} & \textbf{92.3±0.52} & \textbf{93.4±0.36} & \textbf{94.4±0.23} \\ 
\bottomrule
\end{tabular}}
\label{link_prediction}
\end{table*}

\begin{figure*}[htpb]
\centering
\includegraphics[width=0.95\textwidth]{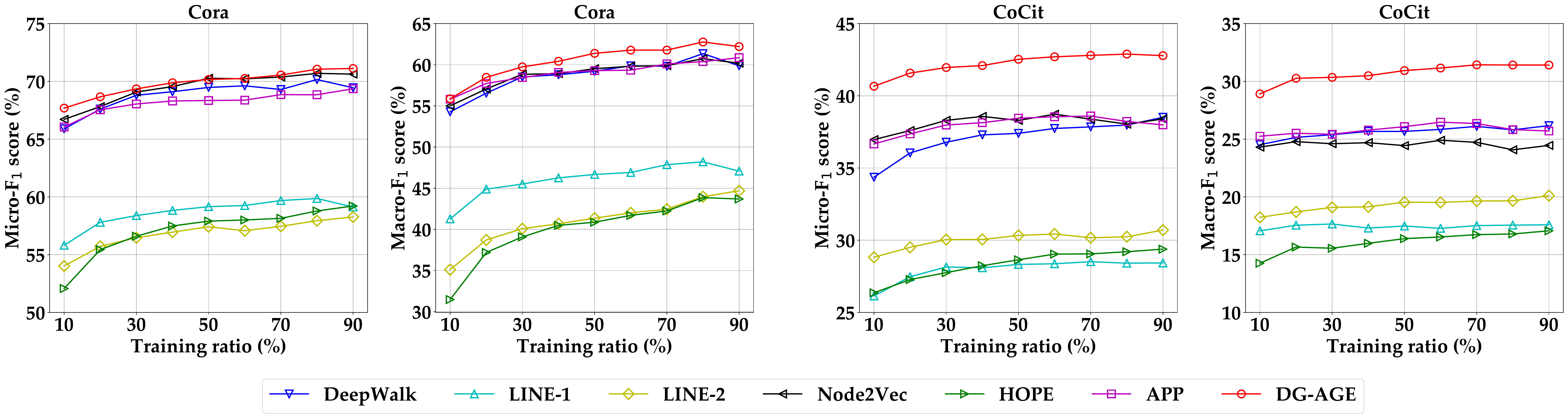}
\vspace{-0.4cm}
\caption{Node classification performance of \DGModel~on \textit{Cora} and \textit{CoCit}. 
The $x$ axis denotes the training ratio of labeled nodes, and the $y$ axis denotes Micro-F1 or Macro-F1 score.
}
\label{figure3}
\end{figure*}

\noindent
\textbf{Link Prediction.}
For the link prediction task, we predict missing edges given a graph with a fraction of removed edges. 
A fraction of edges is removed randomly to serve as test split while the remaining network are utilized for training. 
Specifically, we remove 50\% edges in \textit{Cora}, \textit{Epinions} and \textit{Google}, and 40\% edges in \textit{Twitter}. 
Since we are interested in both the existence and the direction of the edge between two nodes, we reverse a fraction of positive node pairs to replace the original negative samples if the edges are not bi-directional. 
The reversed ratio $\gamma\in(0,1]$ means what fraction of positive edges from the test data are reversed as negative examples and a value of 0 corresponds to the classical undirected graph setting where all the negative edges are sampled from random node pairs. 

We summarize AUC scores of all methods in Table~\ref{link_prediction}. 
Note that some undirected graph embedding methods' performances (e.g., DeepWalk) decrease rapidly with the increase of reversed positive edges because they cannot model the asymmetric proximity. 
The directed graph embedding methods like HOPE and APP show poor performance on \textit{Cora} and \textit{Epinions}. 
The reason is that these methods treat the source role and target role of one node separately, which renders them less robust. 
We can see that \DGModel* shows much better performance than HOPE and APP across datasets. 
This is because the negative samples of \DGModel* are generated directly from a continuous distribution and thus it is not sensitive to different graph structures. 
Moreover, \DGModel~outperforms \DGModel* as it utilizes two generators mutually updating each other for more robust source and target vectors. 
Overall, \DGModel~shows more robustness across datasets and outperforms all baselines on the link prediction task.

\begin{figure}[tbp]
\centering
\includegraphics[width=0.44\textwidth]{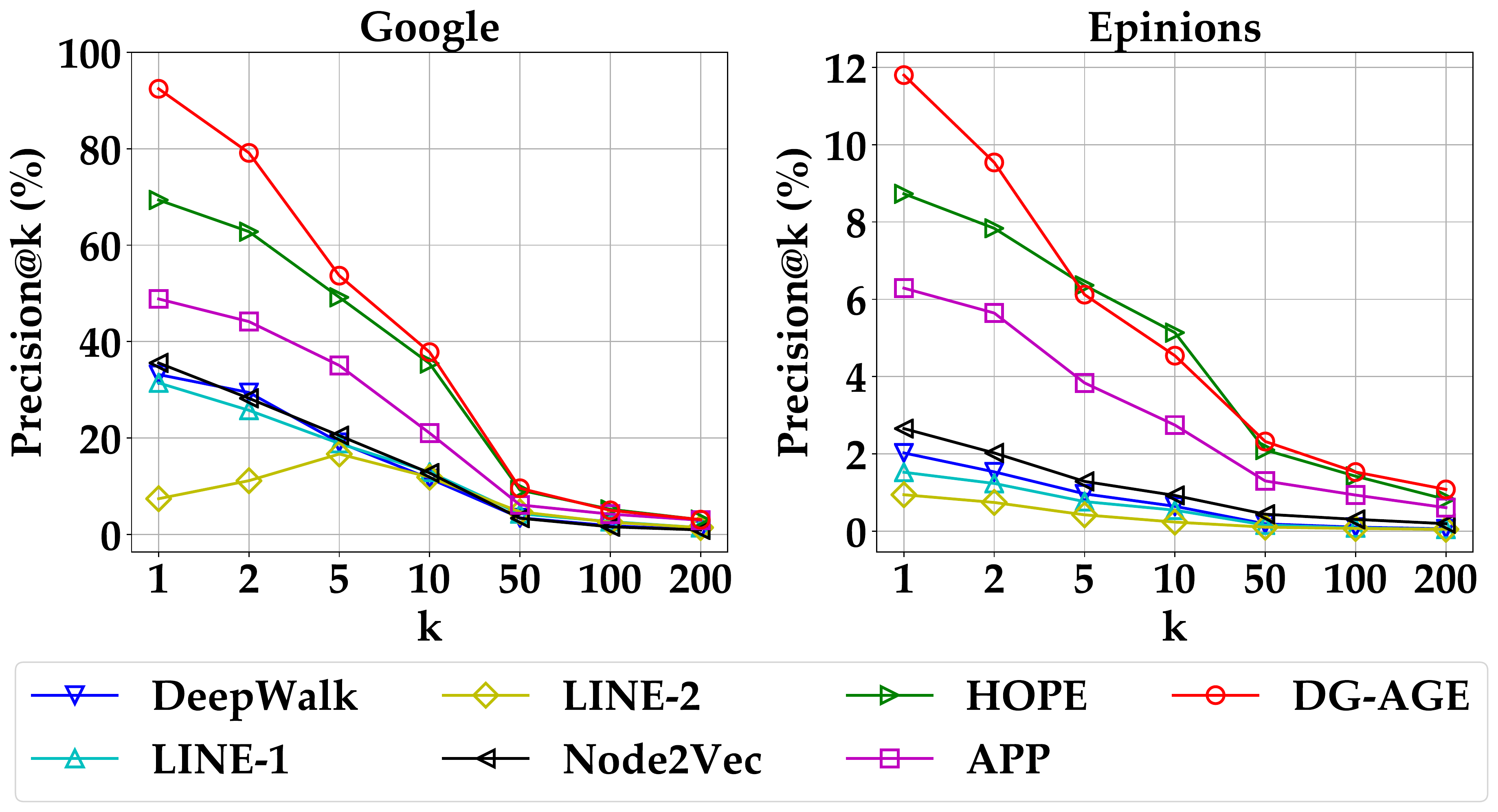}
\caption{Graph reconstruction performance of \DGModel~on \textit{Google} and \textit{Epinions}.}
\label{figure4}
\end{figure}

\noindent
\textbf{Node Classification.}
For the node classification, we evaluate \DGModel~on two directed graph dataset \textit{Cora} and \textit{CoCit}. 
Note that for the methods using both source and target embedding matrices, we set the dimension $d$ of each embeddings to 64 and concatenate the two embedding vectors into a 128-dimensional vector to represent each node. 

Fig.~\ref{figure3} summarizes the experimental results with varying the training ratio of the labeled nodes. 
First, \DGModel~consistently outperforms all baseline methods across all training ratios on both datasets, which demonstrates that \DGModel~can effectively capture the neighborhood information in a robust manner through the adversarial learning framework.
Second, the undirected graph embedding methods DeepWalk and node2vec perform as well as the directed method APP.
This suggests that the directionality might have limited impact on performance for node classification task on these two datasets. 
Third, we notice that HOPE performs poorly on this task, as it is hard to generalize to different tasks with a particular proximity measure. 

\noindent
\textbf{Graph Reconstruction. }
Considering that the direction of edges may directly affect the topology structure of directed graphs, we want to verify the topology learning capability of \DGModel. 
We perform the graph reconstruction task on \textit{Google} and \textit{Epinions} and randomly sample 10\% nodes of each dataset as the test data. 
For the graph reconstruction task, we obtain the $k$-nearest target neighbors with a given $k$ ranked by reconstructed proximity, which is calculated by inner product. 
Then we reconstruct the graph edges based on the proximity between nodes. 

We plot the average precisions corresponding to different values of $k$ in Fig.~\ref{figure4}. 
The results show that \DGModel~mostly outperforms all baselines on both datasets. 
On \textit{Epinions}, HOPE outperforms \DGModel~when $k=5$ and $k=10$. 
It may be because HOPE uses high-order proximity as the weights of directed edges to reconstruct more edges. 
On \textit{Google}, \DGModel~shows an improvement of around 33\% with $k=1$ over the second best method HOPE. 
Some methods (e.g., node2vec) that focus on undirected graphs exhibit good performance in link prediction but show poor performance in graph reconstruction. 
This is because graph reconstruction is harder than link prediction as the model needs to distinguish a small number of positive edges from a large number of negative edges. 
It further proves the benefit of jointly learning the source and target vectors for each node. 

\begin{figure}[t]
\centering
\vspace{-0.25cm}
\subfloat[Sparsity.]{
\label{figure5:a}
\includegraphics[width=0.48\linewidth]{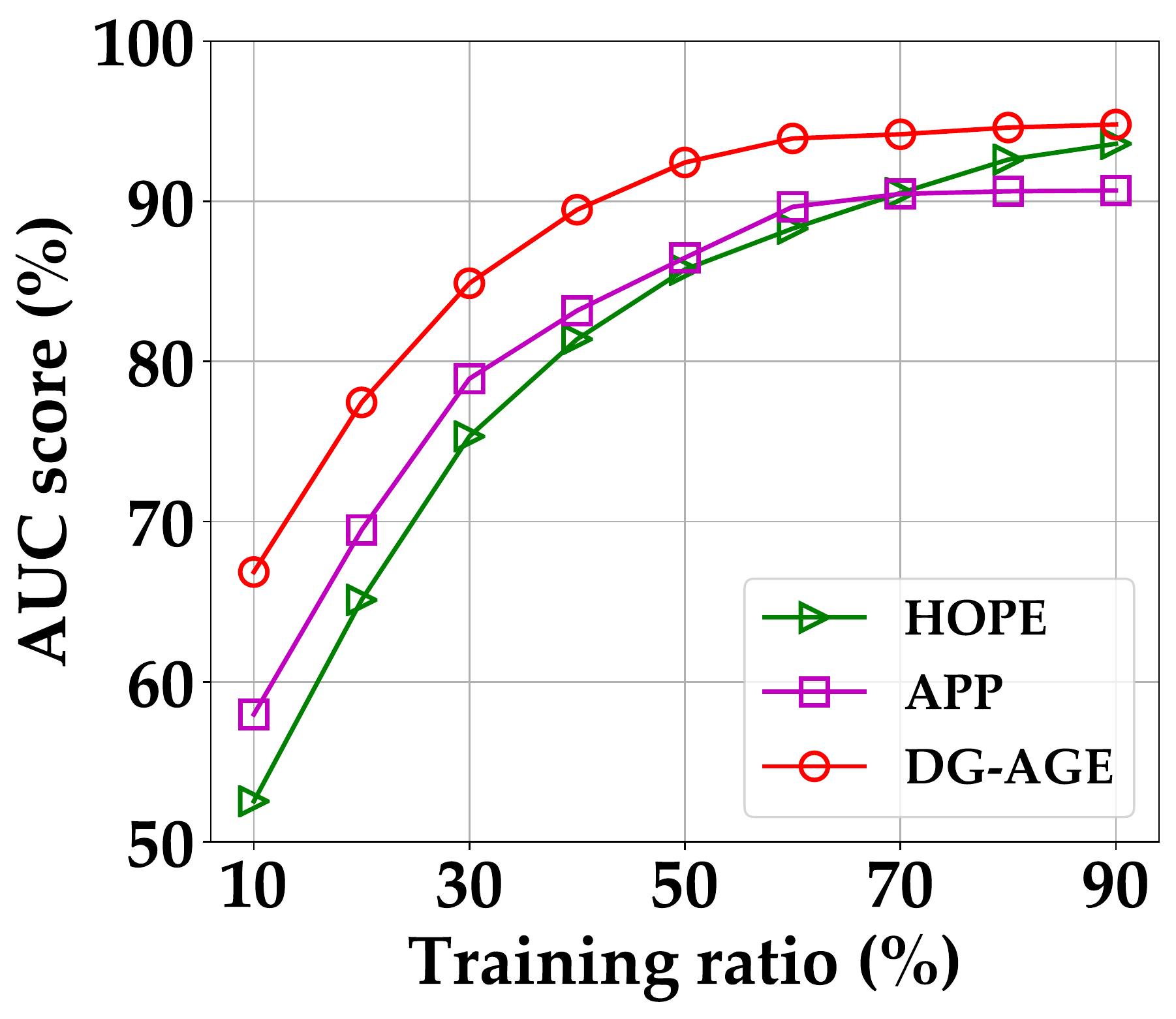}
}
\subfloat[Learning curves.]{
\label{figure5:b}
\includegraphics[width=0.48\linewidth]{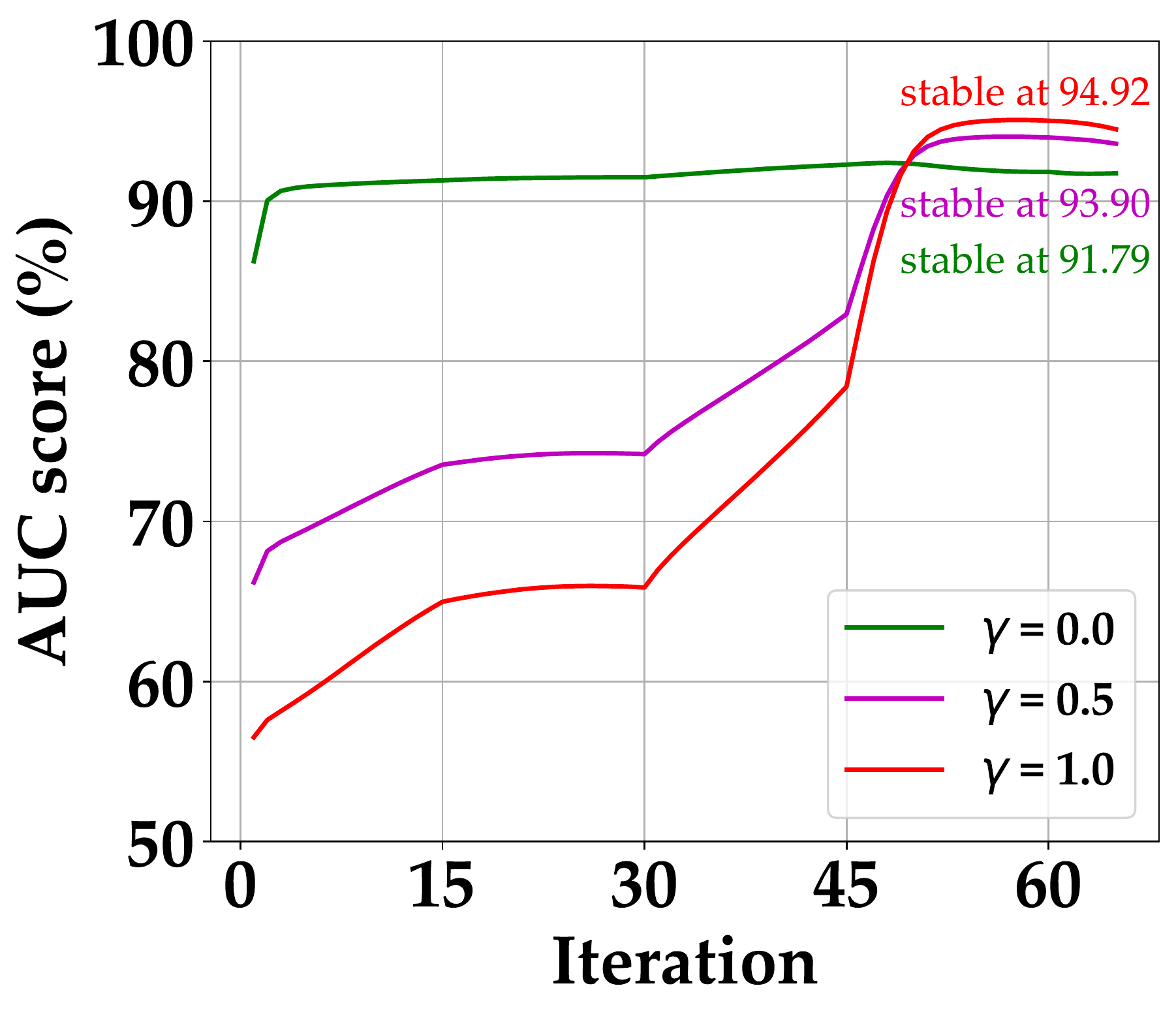}
}
\caption{Link prediction performance and learning curves of \DGModel~on \textit{Google}.}
\label{figure5}
\end{figure}

\noindent
\textbf{Sparsity and Learning Analysis. }
For the directed graph, we analyze the performance of models under different graph sparsity levels and the converging performance of \DGModel.  
We choose \textit{Google} as the evaluation dataset because it is much denser than the others. 
We first investigate how the graph sparsity affects the three directed graph embedding methods HOPE, APP and \DGModel. 
This training setting is the same as in the link prediction task and 50\% positive edges of test set are reversed to form negative edges. 
We randomly select different ratios of edges from the original graph to construct graphs with different sparsity levels.

Fig.~\ref{figure5:a} shows the results with respect to the training ratio of edges on \textit{Google}. 
We can see that \DGModel~consistently and significantly outperforms HOPE and APP across different training ratios. 
Moreover, \DGModel~still achieves much better performance when the network is very sparse while HOPE and APP extremely suffer from nodes with very low outdegree or indegree as mentioned before. 
It demonstrates that the proposed \DGModel, which is designed to jointly learn a node’s source vector and target vector, can significantly improve the representation robustness. 

Next, we investigate performance change with respect to the training iterations of the discriminator $D$. 
Fig.~\ref{figure5:b} shows the converging performance of \DGModel~on \textit{Google} with different ratio of reversed positive edges of test set (results on other datasets show similar trends and are not included here). 
With the increase of iterations of $D$, the performance of \DGModel~with the $\gamma=0$ (i.e. random negative edges in test set) keeps stable first and then slightly increases. 
Besides, the training curve trend of \DGModel~with the $\gamma=1.0$  (i.e. all positive edges except bi-directional edges are reversed to create negative edges in the test set) changes every 15 iterations (i.e. one epoch). 
The training curve trend of \DGModel~with the $\gamma=1.0$ rises gently during second epoch (i.e. from the 16-th iteration to the 30-th iteration) for the generator $G$ which is still been poorly trained at the moment. 
The trend rises steep in the following epochs where $G$ is being able to generate close-to-real fake samples. 

\renewcommand{\multirowsetup}{\centering}
\begin{table*}[!t]
\caption{Summary of link prediction AUC scores (\%), node classification Micro-F1 and Macro-F1 scores (\%) on heterogeneous information networkss. \\
(Result: average score ± standard deviation; \textbf{Bold}: best; \underline{Underline}: runner-up.)}
\centering
{
\scriptsize
\begin{tabular}{c|ccc|ccc|ccc}
\toprule
\multirow{2}{*}{Method} & \multicolumn{3}{c}{DBLP}                                     & \multicolumn{3}{c}{Yelp}                                  & \multicolumn{3}{c}{AMiner}                                                                \\ \cline{2-10} 
                        & AUC            & Micro-F1         & Macro-F1          & AUC            & Micro-F1         & Macro-F1               & AUC            & Micro-F1         & Macro-F1           \\ 
\midrule
DeepWalk        & 56.3±0.72   & 92.0±0.71   & 92.4±0.59   & 78.3±0.22   & 82.6±0.81   & 75.5±0.67   & 51.8±0.81   & 95.2±0.30   & 94.6±0.17       \\
LINE-1          & 72.2±0.30   & 92.4±0.04   & 92.1±0.85   & 79.7±0.95   & 82.3±0.38   & 74.4±0.15   & 64.1±0.58   & 97.8±0.25   & 97.1±0.04       \\
LINE-2          & 65.0±0.38   & 91.4±0.20   & 91.7±0.80   & 67.5±1.12   & 75.9±0.57   & 55.2±0.68   & 51.1±0.11   & 94.7±0.19   & 93.4±0.59       \\
\midrule
GraphGAN        & 53.3±0.62   & 92.0±0.39   & 92.1±0.11   & 76.3±0.57   & 81.0±0.16   & 72.7±0.70   & -   & -    & -         \\
ANE             & 54.3±0.47   & 91.4±0.17   & 91.5±0.47   & 73.3±0.39   & 82.3±0.41   & 76.2±0.92   & 52.8±0.16   & 92.6±0.23    & 92.0±0.13      \\
\midrule
HIN2vec         & 79.5±0.39   & 91.4±0.40   & 91.2±0.48   & 79.6±0.58   & 83.5±0.14   & 76.1±0.50   & 78.7±0.13   & 98.0±0.12    & 97.8±0.05      \\
Metapathvec     & 59.2±0.21   & 92.9±0.34   & 93.0±0.41   & 78.0±0.12   & 79.5±0.86   & 78.8±0.73   & 76.2±0.70   & 98.5±0.28    & 98.6±0.09      \\
TransE          & 76.3±0.07   & 90.2±0.32   & 91.2±1.05   & 77.3±0.57   & 82.5±0.03   & 75.4±0.76   & 75.6±0.21   & 97.1±0.66    & 96.4±0.35      \\
\midrule
\HINModel-TE       & 79.1±0.45   & 91.318   & 92.5±0.13   & 79.9±0.03   & 84.2±0.35   & 79.5±0.30   & 78.65±0.74   & 97.7±0.13    & 97.7±0.10      \\
\HINModel-TH       & \underline{81.3±0.55}   & \underline{94.2±0.58}   & \underline{93.9±0.21}  & \underline{81.3±0.09}   & \underline{86.1±0.17}   & \underline{81.2±0.52}  & \underline{81.5±0.13}   & \underline{98.5±0.08}  & \underline{98.8±0.07}         \\ 
\HINModel-TD       & \textbf{83.2±0.28}   & \textbf{95.2±0.08}   & \textbf{94.1±0.16}  & \textbf{82.1±0.15}   & \textbf{86.6±0.18}   & \textbf{82.1±0.84}  & \textbf{83.1±0.84}   & \textbf{98.7±0.25}   & \textbf{98.9±0.01}      \\
\bottomrule
\end{tabular}
}\label{HGGAN}
\end{table*}

\subsection{Performance Evaluation of \HINModel}
In order to evaluate the performance of \HINModel, we compare it with several methods, including traditional graph embedding methods, GAN-based graph embedding methods and heterogeneous information networks embedding methods. 
We present three versions of \HINModel~(HGGAN-TE, \HINModel-TH and \HINModel-TD) based on TransE, TransH and TransD.
The result does not include the results of GraphGAN on \textit{AMiner}, because it cannot perform on the large dataset. 

\begin{figure}[!t]
\centering
\subfloat[Sparsity.]{
\label{fig:HGGAN-GR:a}
\begin{minipage}{0.48\linewidth}
    \centering
    \includegraphics[width=1\linewidth]{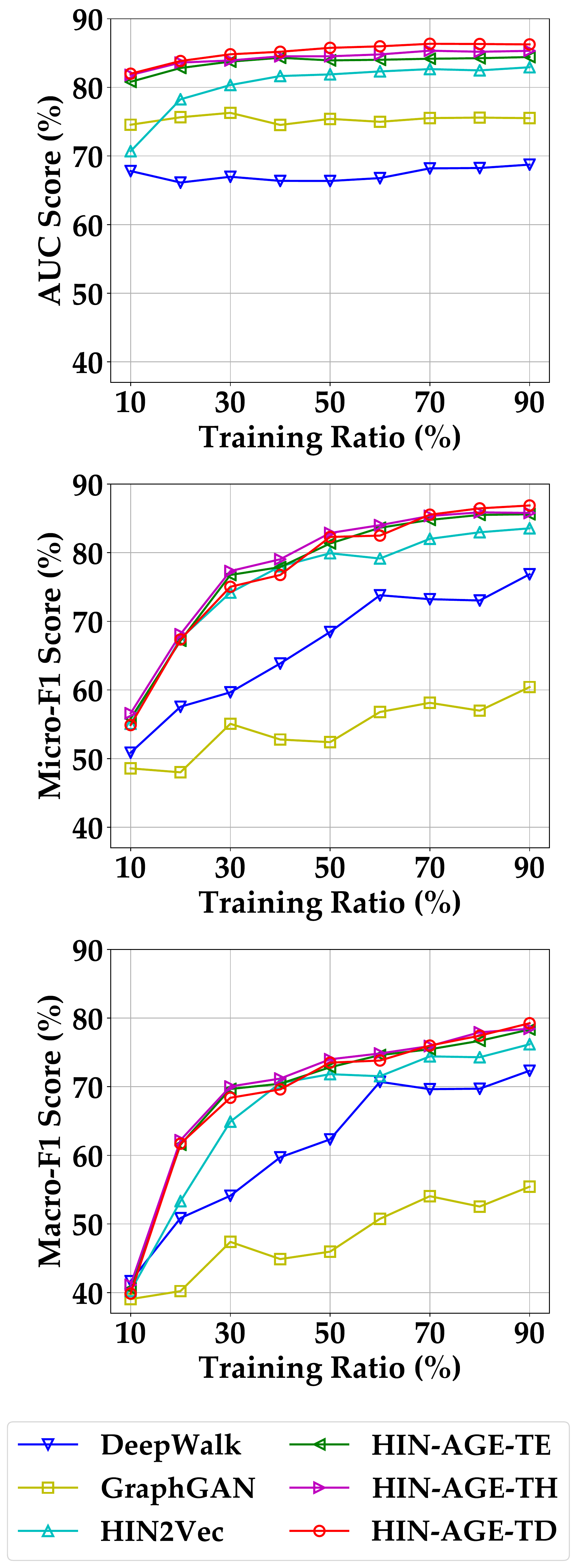}
\end{minipage}

}%
\subfloat[Learning curves.]{
\label{fig:HGGAN-GR:b}
\begin{minipage}{0.48\linewidth}
    \centering
    \includegraphics[width=1\linewidth]{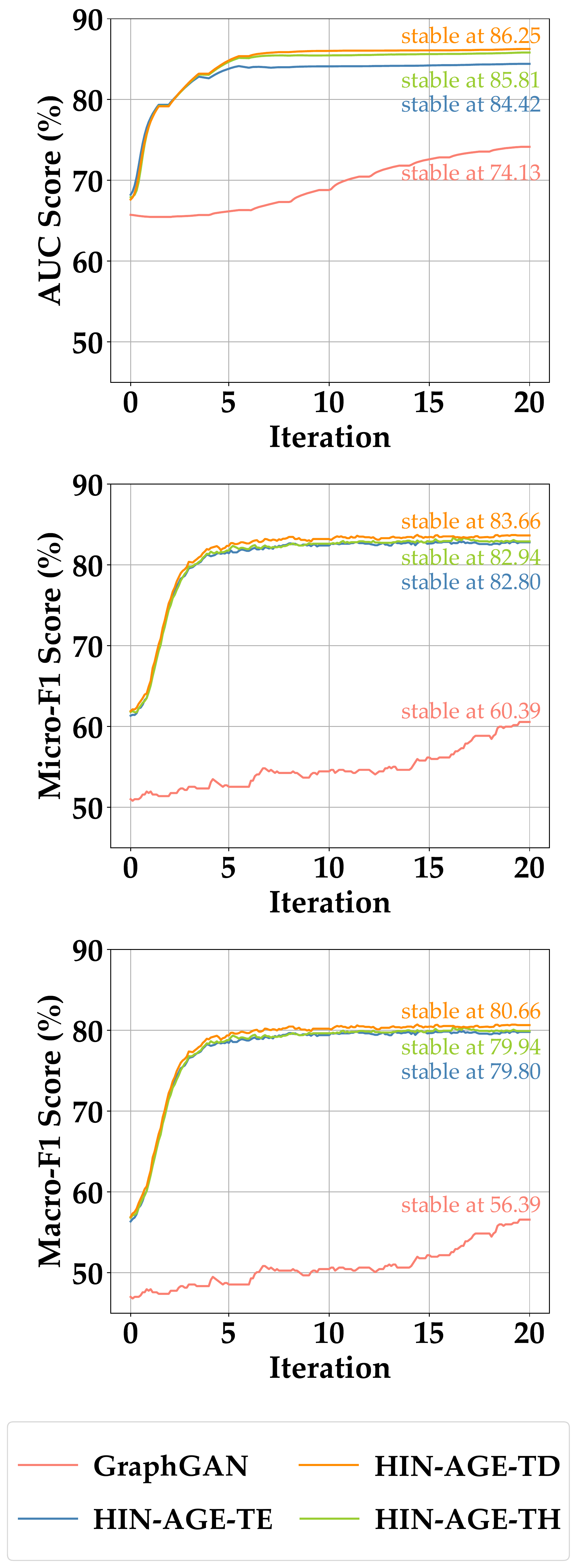}
\end{minipage}
}%
\caption{Performance with different graph sparsity and learning curves of \HINModel~on \textit{Yelp}. }
\label{fig:HGGAN-GR}
\end{figure}

\noindent
\textbf{Performance Analysis. }
For the link prediction task, we predict user-review links in \textit{Yelp} and author-paper links in \textit{DBLP} and \textit{AMiner}. 
For positive samples, we randomly hide 20\% connected node pairs in \textit{Yelp}, \textit{DBLP} and \textit{AMiner} as test set and the remaining 80\% as the training set. 
For the node classification task, we evaluate the proposed \HINModel~and other baseline methods on three heterogeneous information networks datasets \textit{DBLP}, \textit{Yelp} and \textit{AMiner}. 
Similar to the link prediction task, we sample 80\% labeled nodes as the training data and predict the labels for the other 20\% labeled nodes. 

As shown in Table~\ref{HGGAN}, we can observe that \HINModel-TE, HINGAN-TH, and \HINModel-TD always outperform baseline methods including TransE. 
It indicates that the proposed \HINModel~can better preserve the heterogeneous semantics. 
In addition, we observe that \HINModel-TD can achieve higher AUC scores compared with \HINModel-TE and \HINModel-TH. 
It may be because that \HINModel-TE is only suitable for handling one-to-one relationships, not for one-to-many and many-to-one relationships. 
Although TransH can handle one-to-many and many-to-one relationships, it's not suitable for multiple relationships because it embeds entities and relationships into the same semantic space.
Instead of utilizing a single vector to jointly embed nodes and edges, \HINModel-TD uses one vector to represent semantics and another to construct a mapping matrix, which can better retain the semantic information of nodes and edges in heterogeneous information networks.



\noindent
\textbf{Sparsity and Learning Analysis.}
We further evaluate the generalization ability of \HINModel-TE, \HINModel-TH, and \HINModel-TD with different graph sparse conditions on \textit{Yelp}. 
We randomly sample different ratios from 10\% to 90\% of the original training set as training data and randomly sample 10\% of the nodes outside the training set as the test data. 
Then, we analyze the learning process of the three variant models. 

Fig.~\ref{fig:HGGAN-GR} shows the results of link prediction and node classification tasks on \textit{Yelp}. 
Fig.~\ref{fig:HGGAN-GR:a} shows the performance of models with different training ratios. 
It can be observed that \HINModel~consistently outperforms all baselines for both tasks, even when the training ratio is small. 
Fig.~\ref{fig:HGGAN-GR:b} shows the learning curves of \HINModel-TE, \HINModel-TH, and \HINModel-TD, with the results consistently outperforming baselines. 
In addition, we can observe that the learning curves of the three variants of \HINModel~are similar, indicating that the framework is stable in the training process.  

\section{Related Work}
\label{sec:Related Work}
In this section, We first briefly review the graph representation learning methods including the shallow methods and the deep methods. 
Then we review the graph embedding based on generative adversarial network specifically. 

\subsection{Graph Representation Learning} 
Graph representation learning methods can be classified into three categories: matrix factorization based models, random walk based models and deep learning based models. 
The matrix factorization-based models (e.g., GraRep~\cite{cao2015grarep} and M-NMF~\cite{wang2017community}) first preprocess the adjacency matrix which preserves the graph structure, and then decompose the prepocessed matrix to obtain graph representations. 
The random walk based models (e.g.,  DeepWalk~\cite{perozzi2014deepwalk},  LINE~\cite{tang2015line},  PTE~\cite{tang2015pte} and node2vec~\cite{grover2016node2vec})
sample the node sequences to input into Skip-gram model~\cite{mikolov2013distributed} by random walking on the graph and they can be unified into the matrix factorization framework with closed forms~\cite{qiu2018network}. 
However, most of these methods ignore the noise in the real network, and the negative samples used are not strong enough and have poor robustness. 

There are some works focus on learning representations for directed graphs, which commonly learn source embedding and target embedding for each node. 
HOPE~\cite{ou2016asymmetric} derives the node-similarity matrix by approximating high-order proximity measures
and then decomposes the node-similarity matrix to obtain node embeddings. 
APP~\cite{zhou2017scalable} designs an asymmetric proximity preserving directed network embedding approach via random walk with restart, which implicitly preserves the Rooted PageRank score for node pairs. 
NERD~\cite{khosla2018node} generates role-specific node neighbors with a plain alternating random walk strategy and learns node representations in their related source/target nodes. 
ATP~\cite{sun2019atp} incorporates graph hierarchy and reachability to construct the asymmetric matrix. 
For directed graph, most methods are shallow, failing to capture the highly non-linear property in graphs and learn robust node embeddings. 

The graph representation learning models constructed for homogenous graphs are not suitable for heterogenous graphs~\cite{sun2012mining}. 
Recent emerging research directions in heterogeneous information networks embedding can be divided into the following two categories: random walk based models and knowledge graph embedding models. 
The random walk based methods model structural and semantic correlations in heterogeneous information networks simultaneously, such as metapath2vec~\cite{dong2017metapath2vec} and HIN2vec~\cite{fu2017hin2vec}.
These methods design meta-path based random walks or specific random walk strategies to obtain the neighborhood of nodes and perform node embeddings. 
HERec~\cite{shi2018heterogeneous} also designs a meta-path-based random walk strategy and further integrate node embeddings into an extended matrix factorization model. 
The knowledge graph representation learning methods learn low-dimensional embeddings of entities and relations while capture relative semantic meanings~\cite{ji2020survey}. 
However, these methods always need domain knowledge to design meta-paths and walk strategies, and they are difficult to be applied to complex, large-scale heterogeneous information networks. 

\subsection{Generative Adversarial Networks based Graph Embedding}
Recently, Generative Adversarial Network (GAN)~\cite{goodfellow2014generative} attracts increasing attention among researchers due to its impressing performance on the unsupervised tasks. 
GAN can be considered as playing a game-theoretical min-max game between the generator and the discriminator.
Several methods~\cite{wang2018graphgan,dai2018adversarial,yu2018learning,pan2019learning,gao2019progan,sang2019aaane,zhang2019dane,dai2019adversarial} have been proposed to apply GAN for graph embedding to achieve models robustness and generalization. 
GraphGAN~\cite{wang2018graphgan} samples negative nodes in the sampling distribution.
ANE~\cite{dai2018adversarial} regularizes graph embedding learning, whic contains a structure preserving component and an adversarial learning component for obtaining structural properties and robust representations, respectively.
NetRA~\cite{yu2018learning} and ARGA~\cite{pan2019learning} adopt adversarially regularized autoencoders to learn smoothly embeddings.
ProGAN~\cite{gao2019progan} employs triplets of nodes for discovering the complicated latent proximity.
AAANE~\cite{sang2019aaane} designs an attention-based autoencoder to automatically vote for the robust and stable node representations.
DANE~\cite{zhang2019dane} employs graph convolutional network to get transferable node embeddings on different networks.
DWNS~\cite{dai2019adversarial} designs two adversarial training regularization method, including embedding adversarial perturbations with an adaptive $L_2$ norm constraint and enforcing reconstructed adversarial examples.
However, the above methods generate the samples from the original graph, and it cannot learn the unseen information of the graph and difficult to extend to the large-scale network.




\section{CONCLUSIONS AND FUTURE WORKS}
\label{sec:Conclusion}
In this paper, we propose a novel robust and generalized framework for adversarial graph embedding, called AGE. 
Specifically, We design the generator(s) and the discriminator that can preserve complex semantic information of the graph by using the continuous implicit distribution of nodes and the semantic information of the graph. 
The computational complexity of the proposed framework is linearly related to the number of edges in the graph, and can generalized well for various graphs. 
We design three model for three typical graphs by simple modifications, demonstrating the flexibility and generalization of the proposed framework. 
The extensive experimental results on the real-world graph datasets demonstrated that our models consistently and significantly outperform the state-of-the-arts methods in the link prediction, node classification, and graph reconstruction tasks. 

In the future, we plan to explore the proposed methods for more types of semantics graphs (e.g., attribute graphs) to further improve the generalization of our framework. 
Another interesting direction is to fuse our framework with other graph embedding methods deeply for better graph representation capability.

\bibliographystyle{IEEEtran}
\bibliography{AGE}

\end{document}